\documentclass{article}

\usepackage{PRIMEarxivSimple}
\usepackage[utf8]{inputenc}
\usepackage[T1]{fontenc}
\usepackage{amsmath,amssymb,amsfonts}
\usepackage{hyperref}
\usepackage{url}
\usepackage{nicefrac}
\usepackage{microtype}
\usepackage[numbers,sort&compress]{natbib}
\usepackage{fancyhdr}
\usepackage{graphicx}
\usepackage{booktabs}
\usepackage{tabularx}
\usepackage{multirow}
\usepackage{array}
\usepackage{arydshln}
\usepackage{makecell}
\usepackage{enumitem}
\usepackage{longtable}
\usepackage{caption}
\usepackage{pdflscape}
\usepackage{afterpage}

\newcolumntype{Y}{>{\raggedright\arraybackslash}X}
\providecommand{\keywords}[1]{\par\noindent\textbf{Keywords:} #1}
\newcommand{\maintablesetup}{%
    \captionsetup{font=footnotesize,labelfont=bf}%
    \footnotesize
    \setlength{\tabcolsep}{5pt}%
    \renewcommand{\arraystretch}{1.15}%
}
\newcommand{\runningtitle}{Intention Driven Match Phase Identification}
\title{Intention Driven Identification of In-Possession Match Phases\\in Association Football through Temporal Graph Learning}

\author{
Yuesen Li$^{1}$, Daniel Link$^{1,2}$\\[0.3em]
\small $^{1}$Technical University of Munich, Experimental Exercise Science, Munich, Germany\\
\small $^{2}$Technical University of Munich, Munich Data Science Institute, Munich, Germany\\
\small \texttt{yuesen.li@tum.de}\\
}

\date{June 2026}

\begin{document}
\maketitle

\vspace{-0.7em}
\begin{center}
\normalsize
\textbf{Preprint: This manuscript has not been peer reviewed}
\end{center}

\begin{NoHyper}
\begingroup
\renewcommand{\thefootnote}{}
\footnotetext{\textbf{Cite preliminary as:} Li, Y. \& Link, D. (2026). Intention Driven Identification of In-Possession Match Phases in Association Football through Temporal Graph Learning. \textit{arXiv preprint}. Advance online publication. https://doi.org/10.48550/arXiv.2606.09289}
\endgroup
\end{NoHyper}

\begin{abstract}
Understanding tactical organisation in association football requires identifying distinct match phases. Yet in-possession phases are rarely directly observable and are shaped by evolving tactical intentions rather than spatial patterns alone. This study proposes a data-driven framework for identifying in-possession match phases from spatiotemporal tracking data. Seven German Bundesliga matches recorded at 25 Hz with TRACAB were analysed. A hierarchical phase model was defined with three tactical intentions (Invade Opponent Space, Keep Possession, Scoring) and six phases (Build Up, Progression, Counter Attack, Maintenance, Sustained Threat, Finishing). A Temporal Graph Attention Network (T-GAN) was developed to combine frame-level player-interaction graphs, contextual features, and Transformer-based temporal modelling. Performance was evaluated using frame-level F1 for raw classification and temporal Intersection over Union (tIoU) for sequence-level agreement. T-GAN achieved macro-average frame-level F1 scores of 0.87 at the intention level, 0.76 for invasion-related phases, and 0.79 for scoring phases. However, sequence-level evaluation showed that high frame-level F1 could still mask fragmented segments and boundary misalignments. After filtering, mean class-wise tIoU increased from 0.68 to 0.74 for intentions and from 0.55 to 0.61 for phases. Model comparisons indicated that Transformer-based sequence modelling was the main driver of coherent segmentation, whereas graph-based relational modelling was most beneficial for transition-like phases, especially Counter Attack. A further analysis identified several common sources of sequence-level misalignment, mainly involving Build Up/Progression ambiguity, Build Up/Maintenance confusion, Counter Attack/Progression inconsistency, and shortened pre-shot Finishing segments. Overall, the framework translates continuous tracking data into tactically interpretable phase representations for automated annotation, tactical analysis, and playing-style profiling.
\end{abstract}


\section{Introduction}

Recognising human intentions from movement trajectories is an important problem across a wide range of domains \citep{RN617,RN620}. In public safety, intelligent surveillance and healthcare, trajectory-based analysis can support the identification of suspicious or anomalous behaviours in complex environments \citep{RN618,RN622,RN623,RN621}. More broadly, trajectory data have become a key basis for understanding and modelling human behaviour in pedestrian and crowd systems \citep{RN610,RN608,RN609}. Enabled by advances in GPS devices, video tracking, and wearable sensors, large-scale spatiotemporal movement data are now increasingly available, creating substantial opportunities for data-driven intention recognition and behaviour modelling \citep{RN57,RN607}. However, these data also pose substantial challenges as human trajectories are high-dimensional, continuous, and temporally evolving, while intention-driven behaviours often lack explicit boundaries \citep{RN611}. It must instead be inferred as latent, context-dependent segments of variable length within long movement streams \citep{RN57,RN607}.

A setting that provides optimal conditions for studying human intention is sports. Especially elite football provides a particularly valuable context for developing and validating computational approaches to collective human behaviour. Its spatiotemporal dynamics as an invasion team sport \citep{RN57} are complex enough to explore meaningful modelling tasks, yet sufficiently constrained by rules, player roles, and tactical intentions to remain analytically tractable. Football may therefore be regarded as a reduced complexity experimental environment in which coordinated human behaviour can be observed under ecologically valid conditions. This potential has been reinforced by recent advances in performance data collection. High frequency tracking systems and inertial measurement units enable precise measurement of player positions and movements  \citep{RN374}, while event data capture every on-ball action in detail \citep{RN54}. As a result, both the availability and the granularity of football data have increased substantially in recent years \citep{RN377}. Together, these developments have made football an important testbed for computational methods that seek to infer tactical dynamics from continuous spatiotemporal data.

In modern football (association football) analytics, Possession sequences are commonly defined as periods of uninterrupted ball control by a single team. In modern football analytics, they are often treated as a fundamental unit of analysis \citep{RN373,RN584}. Yet a single possession can encompass multiple tactically distinct phases, such as a controlled build up, a rapid counterattack, or a final attacking action. Segmenting match phases is therefore widely considered essential for tactical analysis in football, as it breaks the continuous flow of play into meaningful units associated with different strategic purposes. In practice, coaches have long applied phase-specific principles and used phase-based indexing of match time for training and performance analysis \citep{RN524,RN588,RN586,RN578}. By situating on field events within specific phases, analysts can develop a clearer understanding of tactical context. This, in turn, provides a useful basis for the analysis of additional performance indicators, the development of sports analytics models, and the profiling of playing styles, while also helping to characterise how teams behave both in possession and out of possession \citep{RN594,RN123}. At the same time, defining the characteristics of each match phase remains challenging and, to some extent, subjective, as no universally accepted standard currently exists. Yet classical coaching theory has long suggested that team behaviour is organised primarily around immediate tactical objectives rather than fixed positional templates \citep{RN297,RN578}.

In this study, we focus on the recognition of in-possession match phases in football based on tactical intention. This task is challenging for several reasons. First, conceptual challenge: match phases are not directly observable in either tracking or event data, as their boundaries are rarely signalled explicitly and often emerge gradually through evolving spatiotemporal interactions among players. Second, representation challenge: phase identification depends not only on the spatial arrangement of players at a given moment but also on how these relations evolve over time. Tactically similar situations may therefore belong to different phases depending on the broader sequence context. Third, evaluation challenge: conventional frame-level evaluation is insufficient for this task, because it cannot fully reflect whether predicted segments preserve the temporal coherence of the underlying tactical behaviour. These properties make in-possession match phase identification a particularly demanding problem at the intersection of tactically meaningful segmentation, relational modelling and sequence learning and validation

Against this background, we argue that football in-possession match phases are better characterised as intention-driven temporal processes rather than static spatial configurations. Accordingly, this study makes three main contributions. First, we propose an intention-driven in-possession phase model that aligns tactical definitions with the requirements of sequence-based learning. Second, we introduce a hierarchical Temporal Graph Attention Network (T-GAN) framework that integrates player interaction graphs with Transformer-based temporal modelling to identify match phases from tracking data. Third, we develop sequence-aware evaluation and interpretation methods to assess both segmentation quality and tactical plausibility. The aim of this study is not to establish a universal terminology for all possible match phases. Instead, we operationalise an intention-based model of in-possession phases that can be tested with frame-level tracking data and evaluated as a sequence-segmentation problem.

\section{Related Work}

\subsection{Match Phase Frameworks}

Football match play is commonly described through broad game states that distinguish between possession, non-possession, and transitional moments between attacking and defending \citep{RN123,RN588,RN587}. This basic structure provides an important starting point for organising the continuous flow of play into analytically meaningful units. For example, \citet{RN587} adopted this categorisation in developing an analytical pipeline for player tracking data. However, while the possession--non-possession--transition distinction captures the highest level of game structure, it remains too rough for analysing the tactical organisation of possession play itself.

For this reason, many existing systems introduce further subdivisions within these broad categories. Early work by \citet{RN595} , for instance, distinguished between counter attacks, characterised by rapid progression against a disorganised defence, and elaborate attacks, involving slower and more patient possession, while treating set plays separately. Other researchers and organisations have defined phases using more observable criteria, such as field zones or defensive block height, including distinctions between build up in the defensive third, progression through midfield, final penetration in the attacking third, or, in defensive contexts, high versus low blocks \citep{RN523,RN585,RN594,RN586}. \citet{RN573} classified phases according to their observable tactical form on the pitch and then arranged and combined them into a broader structure. In their framework, open play was divided into attacking and defending phases, with attacking play further separated into possession play and direct play.

These frameworks provide a broad conceptual basis for describing phases of play in Football. However, most organise phases according to observable characteristics such as spatial location, team structure, or attacking pattern, without explicitly accounting for tactical intention. For in-possession analysis, this can create ambiguity because similar observable actions may carry different tactical meanings depending on the surrounding context. For example, controlled ball circulation may reflect Maintenance when the team mainly seeks to retain possession, but may reflect Build Up when the same possession pattern is used to search for attacking opportunities \citep{RN624}. Conversely, the same attacking objective can be realised through different technical-tactical behaviours, such as penetration, offensive support, depth mobility, or width and length \citep{RN625}.

\subsection{Match Phase Recognition}

Previous studies have shown that tactical spatiotemporal patterns in football and other invasion sports can be learned from event and tracking data. These approaches differ in their representational scale. At the action level, \citet{RN596} proposed a pass-embedding framework in which the spatial configuration of players at the moment of a pass was encoded to identify selected phases of play. At the event-sequence level, \citet{RN574} combined event sequences with event-level 360 snapshots to distinguish counterattack-related situations from build-up contexts. At a rough tracking-data level, \citet{RN616} used full positional tracking data to detect tactical states, but aggregated the original 25 Hz data into one-second windows. At the frame level, \citet{RN523} trained a convolutional neural network on expert-annotated optical tracking data to recognise phases of play directly from tracking data. These studies indicate that phase-related tactical patterns are learnable from football data, but they also show that the chosen representational scale strongly constrains the type of tactical information that can be recovered.

A second distinction concerns the formulation of the identification task. Much existing work has focused on binary or selected-pattern recognition, where the aim is to determine whether a specific tactical situation is present. For example, \citet{RN589} distinguished counterattacks from positional attacks in elite handball using graph-based deep learning, while \citet{RN574} focused on identifying counterattack-related scenarios in football. \citet{RN573} further demonstrated the potential of graph-based modelling for hierarchical phase categories, but their supervised model was mainly applied to binary phase classifications (e.g.). These studies provide important evidence that tactical states can be detected from spatiotemporal data. However, in-possession play is not defined by the presence or absence of a single tactical pattern. A possession may evolve from ball retention to build-up, progression, sustained threat, and finishing, while short transition-like behaviours such as counterattacks may occur under specific contextual conditions. Therefore, binary or selected-phase identification does not fully address the problem of segmenting continuous possession sequences into multiple mutually exclusive phases. This requires a multi-class and sequence-aware formulation that can preserve the temporal structure of possession play.

Among existing football studies, \citet{RN523} provide a particularly relevant reference because they demonstrated the feasibility of supervised frame-level phase recognition from optical tracking data and showed clear advantages over a rule-based baseline. However, their framework combines attacking and defensive contexts and was primarily designed to contextualise team formations rather than to segment in-possession sequences according to tactical intention. Moreover, the comparatively limited identification performance for some categories suggests that frame-level input alone does not fully solve the problem of temporally coherent phase segmentation (e.g., F1 = 0.45 for attacking play). Therefore, a remaining challenge is to develop a frame-level, sequence-aware, multi-class framework for recognising multiple in-possession match phases from continuous tracking data.

\subsection{Relational \& Temporal Modelling}

In many football situations, tactically similar states can only be distinguished by considering how players are positioned relative to both teammates and opponents. Previous studies have therefore used graph-based representations to model the relational structure of football scenarios. \citet{RN465} proposed Tactical Graph Networks (TGNets) to capture player interactions, showing in ball-winning outcome prediction that graph-based representations outperformed both state-vector and image-based approaches while requiring substantially lower model complexity. \citet{RN473} used graph-based recurrent neural networks to quantify passing availability, and their results indicated that the derived availability metric was strongly associated with actual pass success and superior to physics-based baselines. \citet{RN592} focused on corner kick situations and built geometric graph neural networks to model player relations and achieve strong predictive performance in receiver and shot prediction tasks, with expert evaluations further confirming that its generated tactical adjustments are realistic and tactically meaningful. In the context of match outcome prediction, \citet{RN593} represented passing interactions as evolving graphs and showed that graph neural networks can predict match outcomes more accurately than conventional feature-based and static network baselines, particularly in in-play, real-time prediction settings.

Related work on match phase detection has likewise begun to explore the potential of graph-based methods. In elite handball, \citet{RN589} trained a graph-based deep learning model to distinguish between counterattacks and positional attacks, using a small set of expert-labelled examples to supervise a GNN that was subsequently applied to hundreds of matches, achieving a balanced accuracy of 86\%. In football, \citet{RN596} introduced a Pass2Vec embedding by using a GNN to encode the spatial configuration of all players at the moment of each pass, thereby generating continuous vector representations of on-ball actions. These low-dimensional embeddings capture contextual characteristics of play and enable similar game situations to cluster together in the learned representation space. \citet{RN573} formalised a comprehensive match phase annotation scheme, constructed a large expert-annotated dataset, and trained a supervised GNN-based model to classify each frame of player-tracking data into binary phase categories. Their approach achieved approximately 80\% accuracy in binary classification of expert-labelled hierarchical phases. As a result, phase recognition requires a representation that preserves the game's relational structure, making graph-based learning a particularly suitable approach.

However, match phase is a concept of time sequence and using only the features or graph structure of a single frame is not sufficient, since the situation before and after it could make a great difference in the phase it belongs to. Sequence-based models, therefore, provide a natural extension for modelling the temporal evolution of such behaviours. Transformer architectures \citep{RN580} have proven particularly effective in modelling long range dependencies and attention-driven sequence dynamics in sports analytics, including but not limited to the match outcome prediction \citep{RN582}, match event prediction \citep{RN581}, trajectory-based playing pattern recognition \citep{RN590}, and sequence-based player performance evaluation \citep{RN579}. In the match phase identification, \citet{RN574} successfully combined Transformers with GNNs to evaluate counterattack decisions in football, using prior events and player positions to identify patterns in successful tactics. Although they investigated only a single type of match phase and the richness of the scenarios was limited by the lack of frame-by-frame tracking data, the study still showed great potential for using a combination of GNNs and Transformers to explore deeper into the task of frame-by-frame match phase identification. To our knowledge, no prior study has used a Transformer-based sequence model to identify match phases directly from frame-level player tracking data.

\section{Phase Model}

Although match phases can be defined for both in-possession and out-of-possession contexts, this study focuses exclusively on in-possession phases. Existing football analysis frameworks commonly distinguish between these contexts as separate tactical problem spaces \citep{RN524,RN123}. In-possession play is particularly suitable for the present study because ball possession provides a clear operational basis for segmenting play into stable sequences and for defining phases relative to the team in control of the ball's objectives \citep{RN373}. Out-of-possession phases involve different tactical principles and would require a separate phase model and annotation scheme. Therefore, focusing on in-possession phases provides a well-defined scope for developing and evaluating the proposed framework.

An in-possession match phase is defined here as a tactically meaningful period within a stable ball possession. To operationalise this idea, we adopt the concept of an Episode, proposed by \citet{RN584}, defined as a stable ball possession lasting at least 3 seconds. Accordingly, we define match phases as periods of stable ball possession with different tactical intentions, or with the same intention but different behaviour in fulfilling it. We define three tactical intentions and 6 phases within them, each corresponding to a different tactical behaviour (see Table~\ref{tab:intention_phase_definition}, and Online Resource 1-).

\subsection{Hierarchical Structure}

Based on this rationale, we organise in-possession behaviour into three tactical intentions: Invade Opponent Space, Keep Possession, and Scoring. These categories represent functionally distinct objectives of possession play: advancing the ball and destabilising the opposition structure, retaining control of possession, and creating or completing immediate scoring opportunities. Together, they capture broad but clearly distinguishable tactical purposes while remaining general enough to accommodate different behavioural realisations.

\begin{figure}[htbp]
    \centering
    \includegraphics[width=0.95\textwidth]{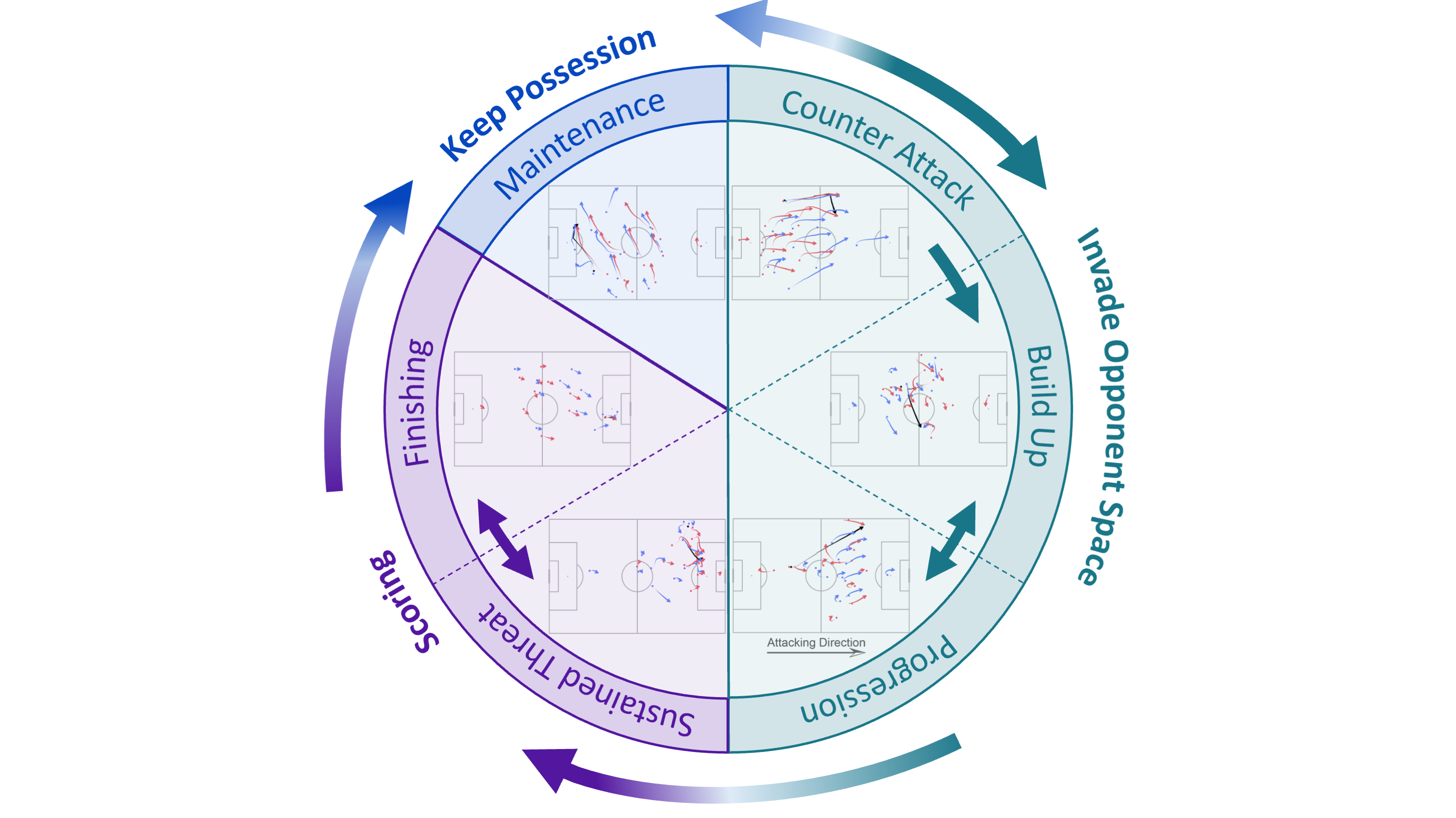}
    \caption{Examples of the six in-possession match phases and their model-defined transition dependencies.}
    \label{figure1}
\end{figure}

\enlargethispage{2\baselineskip}
Within this structure, Invade Opponent Space includes Build Up, Progression, and Counter Attack; Keep Possession is represented by Maintenance; and Scoring includes Sustained Threat and Finishing. Table~\ref{tab:intention_phase_definition} summarises the three intentions and six phases, together with their operational definitions and success criteria. Figure~\ref{figure1} illustrates example situations and the principal transition pathways between phases. Arrows in the outer ring indicate permitted transitions between tactical intentions, while arrows in the inner ring indicate permitted transitions between phase types within the same intention. These rules specify which intentions and phase types can follow one another and guide manual annotation of match phases.

\clearpage
\begin{landscape}
\begin{center}
\centering
\maintablesetup
\captionsetup{hypcap=false}
\captionof{table}{Names and definition of intentions and phases}
\label{tab:intention_phase_definition}
\begin{tabularx}{\linewidth}{p{2.2cm} p{2.2cm} Y Y Y}
\toprule
\textbf{Intention} & \textbf{Phase} & \textbf{Definition} & \multicolumn{2}{c}{\textbf{Rules of success}} \\
\cmidrule(lr){4-5}
 &  &  & \textbf{Successful} & \textbf{Unsuccessful} \\
\midrule

\multirow[t]{3}{2.2cm}{Invade Opponent Space}
& Build Up
& Phase in which a team tries to find low pressure space to make progression, mostly outside the opponent's block.
& Move the ball to a lower pressure area\newline
Find a chance to make Progression\newline
Being fouled
& Lost possession\newline
Be forced to Maintenance \\

& Progression
& A team moves the ball forward to at least the next third in any possible way, usually happens in the middle or the back of the opponent's block.
& Move the ball to at least the next third
& Lost possession\newline
Unsuccessful pass\newline
Be forced back \\

& Counter Attack
& A team initiates attack with high speed immediately after regains possession. Usually when there is no organized defence.
& Successfully move to Finishing or Sustained Threat
& Lost possession\newline
Be interrupted by tactical foul or out of pitch by a tackle\newline
Be forced to Build up or Maintenance \\

\hdashline

Keep Possession
& Maintenance
& The in-possession team is trying to keep their possession, whether willingly or forced.
& Successfully break the opponents' pressure\newline
Successfully move to another phase
& Lost possession\newline
Be forced to make an unsuccessful long ball \\

\hdashline

\multirow[t]{2}{2.2cm}{Scoring}
& Sustained Threat
& The attacking team is moving in the final third, around or at the edge of the opponent's box and trying to find space to create a dangerous situation.
& Successfully move to Finishing\newline
Successfully caused other threatening events, e.g., front freekick, penalty or corner kick
& Lost possession\newline
Be cleared by opponent\newline
Be forced to Maintenance \\

& Finishing
& The final actions which intent to prepare a shot or score directly.
& Goal
& Not goal \\

\bottomrule
\end{tabularx}
\end{center}
\end{landscape}
\clearpage

For example, transitions from Sustained Threat back to Build Up are expected to involve Maintenance, while entry into Sustained Threat is typically preceded by Progression or Counter Attack.

\subsection{Conceptual Advantages}

The proposed phase model should be understood as a more specific extension of in-possession analysis rather than a direct replacement for existing match-phase frameworks. For example, \citet{RN573} proposed a validated and compact annotation scheme for tactical periodization, in which match phases are derived from combinations of game states and team states. This design is well-suited for broad match contextualization because it aims to cover the whole match with a minimal number of semantically non-overlapping classes. However, its offensive open-play categories mainly distinguish between possession play and direct play. While this distinction is useful for general contextualization, it provides limited resolution for analysing how possession sequences evolve internally, for example, from ball retention to build-up, progression, sustained threat, or finishing.

The present study, therefore, adopts a narrower but more fine-grained perspective. By restricting its scope to stable in-possession sequences, the proposed model can define phases based on the team's tactical objective in possession. The first level distinguishes whether the team primarily aims to invade opponent space, retain possession, or create scoring opportunities. The second level then separates the specific behavioural patterns through which these objectives are realised. This hierarchical structure is intended to reduce ambiguity between visually similar possession behaviours, such as ball circulation for retention and ball circulation as preparation for progression, while preserving the temporal organisation of possession play. From a modelling perspective, this decomposition also creates more homogeneous classification tasks, which is beneficial for sequence learning models dealing with complex spatiotemporal behaviours \citep{RN597}.

\section{Phase Identification}

\subsection{General Pipeline}

In this study, in-possession match-phase recognition is formulated as a hierarchical sequence-segmentation problem driven by tactical intention, rather than a flat frame-wise classification task. Tactical intentions serve as a higher-level organisational principle for structuring phases, enabling semantically related behaviours to be grouped while reducing ambiguity between visually similar but strategically distinct situations.

The identification process consists of two stages and is formulated as a hierarchical classification problem with mutually exclusive labels at each level, rather than as a set of independent one-vs-rest phase detectors. At each frame, a possession can belong to only one tactical intention and one corresponding phase. First, tactical intentions are inferred for each stable ball-possession sequence using a Temporal Graph Attention Network (T-GAN), trained on spatiotemporal features computed at the frame level. This intention model performs a three-class classification among Invade Opponent Space, Keep Possession, and Scoring. Second, phase classification is performed within each intention category using dedicated T-GAN models. Specifically, separate models are trained for intention recognition, invasion-related phases, and scoring-related phases, yielding three classifiers: an intention classifier, an invade-opponent-space classifier, and a scoring-phase classifier. The invade-opponent-space classifier distinguishes between Build Up, Progression, and Counter Attack, whereas the scoring-phase classifier distinguishes between Sustained Threat and Finishing. The preliminary predictions are subsequently refined through a filtering procedure and a rule-based correction module grounded in domain specific football knowledge.

\begin{figure}[htbp]
    \centering
    \includegraphics[width=0.95\textwidth]{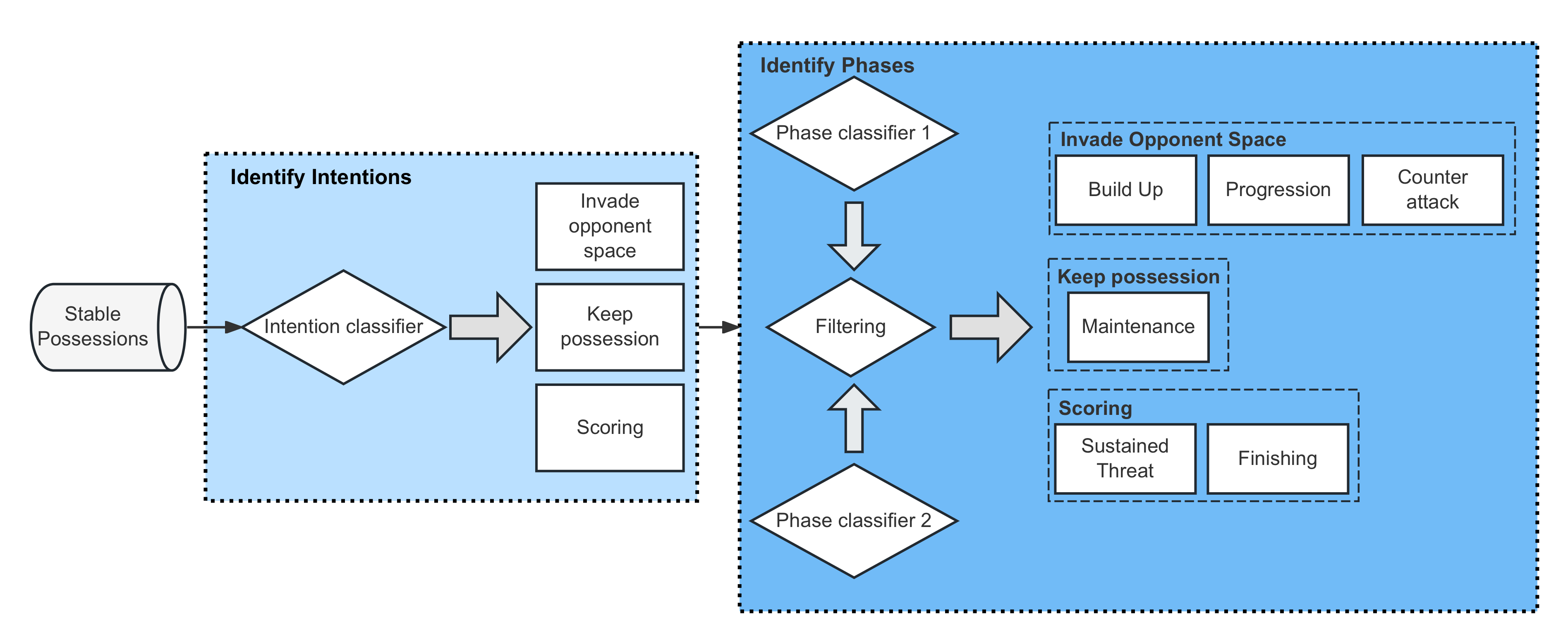}
    \caption{Two-stage phase identification framework. Stable possession sequences are first classified into high-level tactical intentions (invade opponent space, keep possession, and scoring). Based on these intentions, phase classifiers further distinguish finer-grained in-possession phases, with a rule-based fixer applied to ensure temporal consistency and valid phase transitions.}
    \label{figure2}
\end{figure}

Graph \& Feature Representation

In addition to graph-derived features, frame-level global features are incorporated, including action, spatial context, and possession context information. Action features are aligned with the specific frames in which the corresponding events occur. This representation enables T-GAN to combine local player interactions with broader spatial, possession, and event context. The complete feature set, including feature definitions and data types, is provided in Appendix Table~\ref{tab:appendix-feature-set}.

\subsection{Temporal Graph Attention Network (T-GAN) Model}

\begin{figure}[htbp]
    \centering
    \includegraphics[width=0.95\textwidth]{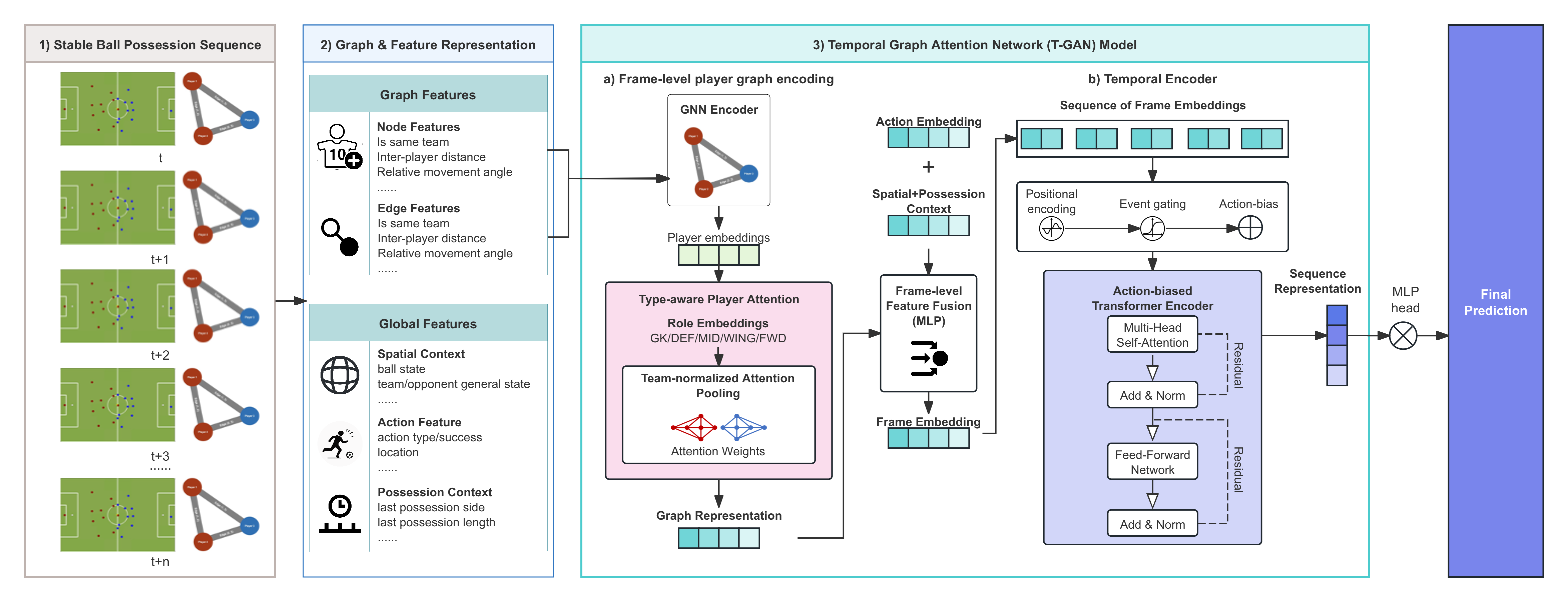}
    \caption{Overview of the proposed Temporal Graph Attention Network (T-GAN) architecture for phase recognition. Frame-level spatiotemporal inputs are first encoded as player-interaction graphs, where node and edge features capture individual states and pairwise relationships. These graphs are processed by GNN layers to produce interaction-aware player embeddings, which are further integrated with global contextual features. A type-aware player attention pooling mechanism aggregates player representations into a unified frame-level embedding. The resulting sequence of embeddings is then passed through a Transformer encoder to model temporal dependencies across frames. Finally, task-specific heads generate predictions for tactical intentions and match phases at both frame and sequence levels.}
    \label{figure3}
\end{figure}

The T-GAN model integrates graph neural networks (GNNs) with attention mechanisms to model player interactions and their temporal evolution during football possessions. As illustrated in Figure~\ref{figure3}, node-, edge-, and global-level features are first organised into frame-wise player interaction graphs. Each graph is then processed by stacked GNN layers to capture intra-frame relational structures among players.

For each frame, the GNN produces an embedding for every player. These player embeddings are then combined into a single frame-level representation via attention pooling. The model learns one attention weight for each player, normalised separately within the attacking and defending teams. Players with higher attention weights therefore contribute more to the final frame representation. To make this aggregation more structured, each player is also assigned a positional role, such as goalkeeper, defender, midfielder, winger, or forward. The role information is used when computing the attention weights. The learned attention weights are retained as supporting outputs and are later summarised by positional group in the Player Attention analysis section.

After the frame-level representations are obtained, they are passed into a Transformer encoder to model the temporal context of the possession sequence. This allows each frame prediction to use information from the surrounding possession sequence, rather than relying only on the current frame. To incorporate on-ball actions, action features are encoded and combined with the frame representations. In addition, an action-dependent bias is added to the Transformer attention scores. This encourages the model to attend more strongly to frames containing explicit actions, such as passes, ball carries, or shots, because these frames often mark important changes in the tactical development of a possession.

The final sequence representation is passed to a fully connected classification layer (multi-layer perceptron). For binary tasks (scoring phases), a sigmoid activation produces probability estimates, whereas for multi-class tasks (intention classification and invasion phases), a softmax activation is used. The predicted class corresponds to the category with the highest probability.

\subsection{Filter}

Filtering was applied after raw prediction to consolidate short isolated label changes. Short predicted runs were reassigned to the local modal label within the model's possession context. The phase-level threshold was 10 frames for Finishing and 50 frames for the other phases. Intention-level thresholds were 50 frames for Invade Opponent Space and Keep Possession, and 25 frames for Scoring.

\section{Experimental Design}

\subsection{Dataset}

Our dataset consists of seven German Bundesliga matches captured by the validated TRACAB optical tracking system \citep{RN374}. The TRACAB system provides positional data for all players and the ball at 25 frames per second, offering sufficiently fine temporal resolution to capture micro-movements, ball dynamics, and rapid tactical shifts within each possession.

Tactical intentions and match phases were manually annotated by the first author, who has formal academic training in sport science and football analytics and the proposed phase definitions. The annotation followed the operational definitions and transition principles described in the phase model, and stable possession episodes were pre-calculated according to the procedures of \citet{RN373} and \citet{RN584}. During labelling, each possession sequence was reviewed using synchronised event and tracking information, and labels were assigned frame-wise based on the tactical objective and the team's behavioural characteristics in possession. The total number of frames and sequences of stable possessions, intentions and phases is shown in Table~\ref{table2}.

\begin{table}[htbp]
\centering
\maintablesetup
\caption{Distribution of labelled possession sequences across the proposed intention-based framework in all 7 matches}
\label{table2}
\begin{tabular}{llrr}
\toprule
\textbf{Sequence category} & \textbf{} & \textbf{Number of Frames} & \textbf{Number of Sequences} \\
\midrule

Stable ball Possession &  & 465233 & 1257 \\

\hdashline

\multirow[t]{3}{*}{Intentions}
& Invade Opponent Space & 303405 & 348 \\
& Keep Possession       & 99742  & 296 \\
& Scoring               & 62086  & 265 \\

\hdashline

\multirow[t]{6}{*}{Phases}
& Build Up          & 178588 & 424 \\
& Progression       & 111976 & 451 \\
& Counter Attack    & 14624  & 67  \\
& Maintenance       & 99742  & 296 \\
& Sustained Threat  & 57745  & 283 \\
& Finishing         & 2558   & 95  \\

\bottomrule
\end{tabular}
\end{table}

\subsection{Evaluation Metrics}

Frame level F1 score was used at the frame level during training and parameter selection, reflecting the model's raw classification performance before filtering and rule-based correction.

At the sequence level, temporal Intersection over Union (tIoU) was used to measure temporal agreement between predicted and observed match phases. Previous studies on in-game status detection in football \citep{RN492} and badminton \citep{RN598} used IoU directly to evaluate their models at the sequence level. Each phase was represented as a segment defined by its start and end frames. For a ground-truth segment (g) and a predicted segment (p), tIoU was calculated as:

\begin{equation}
\operatorname{tIoU}(g,p)=\frac{|g\cap p|}{|g\cup p|}.
\end{equation}

where $g \cap p$ is the temporal overlap and $g \cup p$ is the temporal union.

\begin{figure}[htbp]
    \centering
    \includegraphics[width=0.90\textwidth]{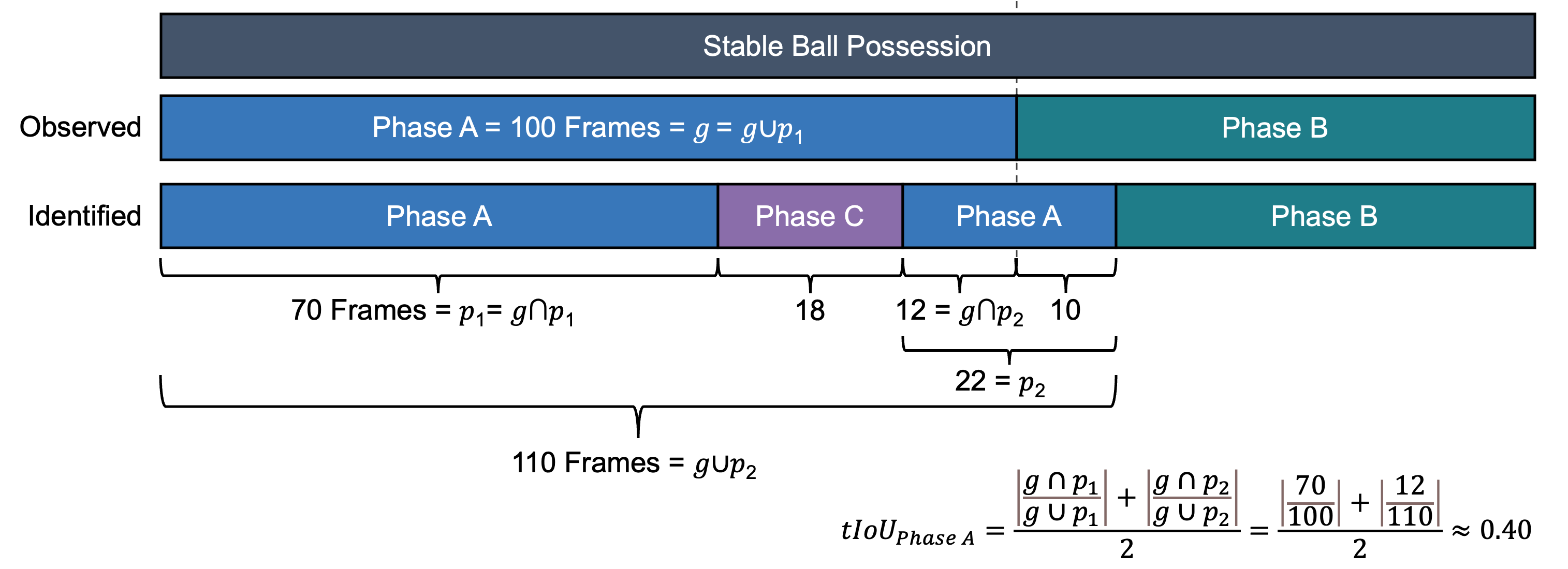}
    \caption{Example of sequence-level temporal Intersection over Union (tIoU) under the strict all-overlap strategy. The upper row shows the observed phase sequence and the lower row shows the predicted sequence. tIoU is computed as the temporal intersection over union for each overlapping segment pair. In this example, the observed Phase A spans 100 frames but is fragmented in the prediction by an intervening Phase C segment, resulting in a lower tIoU (0.40). In this case, Phase C will be filtered in the final prediction and increasing the tIoU to 0.90.}
    \label{figure4}
\end{figure}

A tIoU threshold of 0.5 is often used to describe sufficient temporal overlap in action localization \citep{RN626,RN627}. For each observed segment, tIoU values were averaged across every overlapping predicted segment of the same class. An observed segment without a same-class overlap received zero. Class-wise scores were averaged within each match, then across the seven matches with equal weight. Evaluation preserved possession boundaries and gaps between observed frames. Figure~\ref{figure4} illustrates how the all-overlap calculation penalises fragmented predictions.

\subsection{Baseline Models}

Four baseline models were trained using the same experimental procedure and setup to validate the advantage of our T-GAN model. They are:

\begin{itemize}
\item Rule-based: A rule-based model based on pitch area and ball moving direction
\item Random Forest: a frame-level model using a Random Forest Classifier based on all Global-level features. All features in the Frame global level were used.
\item Pure GNN: a frame-level GNN model using Graph features. The structure remains the same as it is in the T-GAN model. The Frame's global-level features are fed directly into the GNN structures.
\item Pure Transformer: sequence-level Transformer model using Global-level features. The structure remains the same as it is in the T-GAN model. Action-biased attention is used as well.
\end{itemize}

\subsection{Model parameters and training}

Except for the rule-based model, all other models were trained using a 7-fold Leave-one-match-out strategy. In each fold, all possession sequences from one match were used as the test set, while the remaining six matches were used for model training. This procedure was repeated seven times so that each match served as the test set once. The split was performed at the match level to avoid information leakage between training and test data. In each fold, all possession sequences from one complete match were held out as the test set, while the remaining six matches were used for model development. No frames or possession sequences from the held-out test match were used during model training. The Neural Network-based models (Pure GNN, Pure Transformer, and T-GAN) were optimised using the Adam optimiser, and we used a weighted Focal loss to mitigate the impact of an unbalanced data distribution. Other details of the parameter settings and training strategies for the T-GAN model and each baseline model are presented in Appendix Table~\ref{tab:appendix-training-strategy}.

\section{Model Performance \& Comparison}

\subsection{Frame Level Performance}

The frame-level performance of the proposed T-GAN model is summarised in Table~\ref{table3}. Overall, T-GAN achieves consistently strong performance across both intention-level and phase-level tasks, indicating its ability to capture key patterns of in-possession play.

At the intention level, the model reliably distinguishes between the three tactical objectives, suggesting that high-level intentions can be effectively inferred from frame-wise spatiotemporal features. At the phase level, T-GAN shows stable performance for structured phases such as build-up and progression, while comparatively lower performance is observed for more dynamic or short-duration phases, such as counter-attacks and finishing.

Overall, the results indicate that T-GAN provides robust frame level classification performance, with variations largely reflecting the inherent structural differences across phases.

\afterpage{%
\clearpage
\begin{landscape}
\begin{table}[p]
\centering
\maintablesetup
\caption{The performance of each model on frame level during the training process. The F1 Score was calculated as mean$\pm$std value of all 7 folds.}
\label{table3}

\resizebox{\linewidth}{!}{%
\begin{tabular}{lccccccccccc}
\toprule
\multirow{2}{*}{\textbf{Model}}
& \multicolumn{4}{c}{\textbf{Intention}}
& \multicolumn{4}{c}{\textbf{Phase-Invade opponent space}}
& \multicolumn{3}{c}{\textbf{Phase-Scoring}} \\
\cmidrule(lr){2-5}
\cmidrule(lr){6-9}
\cmidrule(lr){10-12}

& \makecell{\textbf{Invade Opponent}\\\textbf{Space}}
& \makecell{\textbf{Keep}\\\textbf{Possession}}
& \textbf{Scoring}
& \makecell{\textbf{Macro-}\\\textbf{average}}
& \textbf{Build Up}
& \textbf{Progression}
& \makecell{\textbf{Counter}\\\textbf{Attack}}
& \makecell{\textbf{Macro-}\\\textbf{average}}
& \makecell{\textbf{Sustained}\\\textbf{Threat}}
& \textbf{Finishing}
& \makecell{\textbf{Macro-}\\\textbf{average}} \\
\midrule

Rule-based model
& $0.76\pm0.02$ & $0.64\pm0.04$ & $0.74\pm0.04$ & $0.71\pm0.03$
& $0.65\pm0.02$ & $0.59\pm0.02$ & $0.24\pm0.14$ & $0.49\pm0.05$
& $0.90\pm0.03$ & $0.28\pm0.09$ & $0.59\pm0.05$ \\

Random Forest
& $0.86\pm0.03$ & $0.71\pm0.06$ & $0.83\pm0.04$ & $0.80\pm0.04$
& $0.85\pm0.06$ & $0.67\pm0.05$ & $0.61\pm0.07$ & $0.71\pm0.05$
& $0.97\pm0.02$ & $0.21\pm0.08$ & $0.59\pm0.05$ \\

Pure GNN
& $0.87\pm0.02$ & $0.76\pm0.04$ & $0.84\pm0.04$ & $0.82\pm0.02$
& $0.82\pm0.05$ & $0.72\pm0.05$ & $0.56\pm0.10$ & $0.70\pm0.03$
& $0.97\pm0.01$ & $0.47\pm0.12$ & $0.72\pm0.06$ \\

Pure Transformer
& $0.90\pm0.02$ & $0.81\pm0.04$ & $0.86\pm0.03$ & $0.86\pm0.03$
& $\boldsymbol{0.86\pm0.04}$ & $0.76\pm0.05$ & $0.56\pm0.12$ & $0.72\pm0.04$
& $\boldsymbol{0.98\pm0.01}$ & $\boldsymbol{0.63\pm0.12}$ & $\boldsymbol{0.81\pm0.06}$ \\

T-GAN
& $\boldsymbol{0.91\pm0.02}$ & $\boldsymbol{0.82\pm0.04}$ & $\boldsymbol{0.87\pm0.03}$ & $\boldsymbol{0.87\pm0.03}$
& $0.85\pm0.04$ & $\boldsymbol{0.76\pm0.04}$ & $\boldsymbol{0.66\pm0.05}$ & $\boldsymbol{0.76\pm0.02}$
& $\boldsymbol{0.98\pm0.01}$ & $0.60\pm0.11$ & $0.79\pm0.06$ \\

\bottomrule
\end{tabular}%
}
\end{table}
\end{landscape}
\clearpage
}

\subsection{Sequence Level Performance}

\begin{figure}[htbp]
    \centering
    \includegraphics[width=0.95\textwidth]{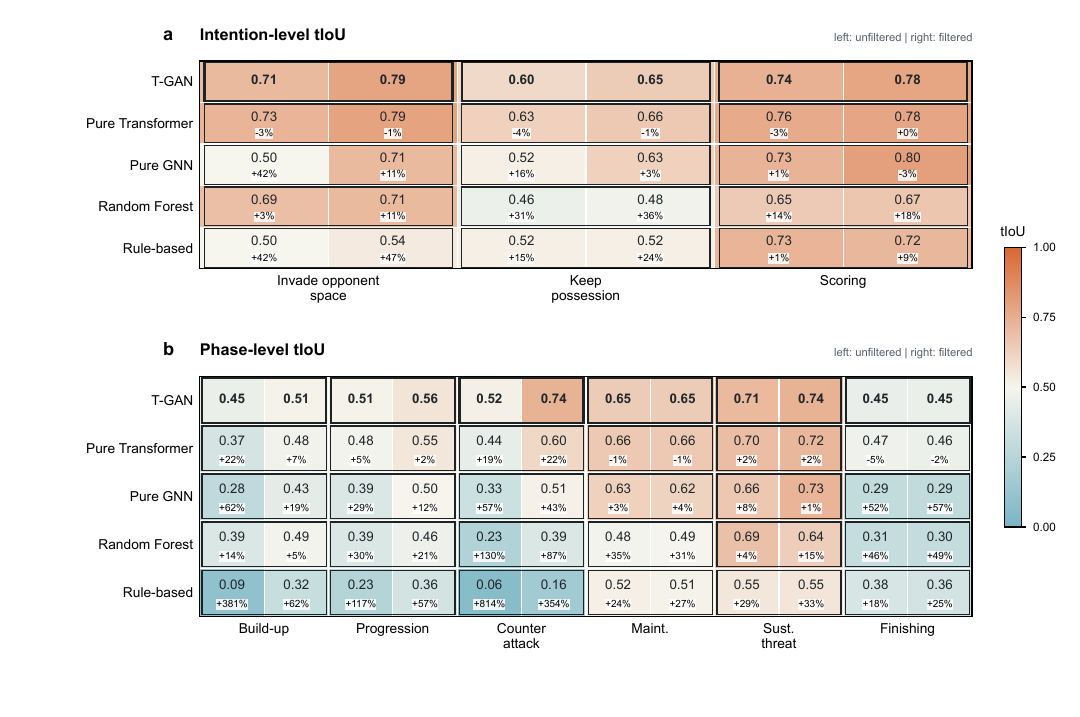}
    \caption{Sequence-level temporal Intersection over Union (tIoU) for all models at the intention level (a) and phase level (b), before and after post-processing. Each cell shows unfiltered results (left) and filtered results (right). Values represent mean tIoU across overlapping predicted and observed segments. T-GAN is the reference model, and percentages indicate differences from it. Higher tIoU indicates better temporal alignment. Filtered results reflect the impact of temporal filtering. Maint. = Maintenance; Sust. threat = Sustained Threat.}
    \label{figure5}
\end{figure}

Figure~\ref{figure5} presents intention-level and phase-level tIoU before and after filtering. Frame-level F1 measures label agreement, whereas tIoU additionally reflects the temporal extent of matching segments. For T-GAN, intention-level macro-average tIoU increased from 0.68 to 0.74, and phase-level tIoU increased from 0.55 to 0.61. These results show an overall benefit of filtering, with changes varying across classes.

At the intention level, T-GAN achieved unfiltered tIoU values of 0.71 for Invade Opponent Space, 0.60 for Keep Possession, and 0.74 for Scoring. After filtering, the corresponding values were 0.79, 0.65, and 0.78. Filtering therefore improved agreement across all three intentions, although the largest increase occurred for Invade Opponent Space.

At the phase level, unfiltered T-GAN scores were 0.45 for Build Up, 0.51 for Progression, and 0.52 for Counter Attack. These scores indicate that correct frame-wise classification did not always recover the complete temporal extent of each observed phase.

After filtering, Build Up increased to 0.51, Progression to 0.56, and Counter Attack to 0.74. Counter Attack showed the largest improvement, consistent with the consolidation of brief interruptions in its predicted sequence. However, this overall benefit did not extend equally to every phase.

Maintenance remained approximately unchanged at 0.65, while Sustained Threat increased from 0.71 to 0.74. Finishing remained more difficult, with tIoU approximately unchanged at 0.45 after the 10-frame filter. Its short duration makes the score sensitive to small boundary deviations. Filtering alone therefore provided little improvement for this phase.

Overall, frame-level accuracy alone did not fully describe sequence agreement. Temporal filtering improved T-GAN's macro-average tIoU, particularly through better Counter Attack segmentation. The effects were class dependent, and coherent segment recovery remained challenging for short Finishing phases.

\subsection{Model Comparison}

Figure~\ref{figure5} compares the models at both levels of the hierarchy. After filtering, phase-level macro-average tIoU was 0.61 for T-GAN, 0.58 for Pure Transformer, and 0.51 for Pure GNN. Random Forest reached 0.46, and the corrected rule-based model reached 0.38. The comparisons support complementary roles for temporal context and relational features, with the clearest graph-related advantage occurring in Counter Attack.

At the intention level, T-GAN and Pure Transformer remained close after filtering, both reaching a macro-average tIoU of 0.74. Random Forest reached 0.62 overall. Pure GNN reached 0.71 overall and achieved the highest Scoring score among these three models, at 0.80. The corrected rule-based scores were 0.54 for Invade Opponent Space, 0.52 for Keep Possession, and 0.72 for Scoring. Its macro-average increased only slightly, from 0.58 before filtering to 0.59 afterwards, and remained below T-GAN across all three intentions.

For Build Up and Progression, T-GAN and Pure Transformer performed similarly after filtering. T-GAN reached 0.51 and 0.56, compared with 0.48 and 0.55 for Pure Transformer. Pure GNN reached 0.43 and 0.50, while the corrected rule-based model reached 0.32 and 0.36. This pattern is consistent with temporal continuity contributing to the recognition of sustained phases. Previous studies have also emphasised sequential decisions and evolving spatial interactions in football \citep{RN174,RN603,RN604}. The present comparisons support this interpretation without establishing that any single representation is necessary for every phase.

Sustained Threat showed relatively high agreement across the learned models. After filtering, T-GAN reached 0.74, Pure GNN 0.73, and Pure Transformer 0.72. Their similar scores indicate that this phase can be recognised from several forms of spatial and temporal information. The corrected rule-based model reached 0.55.

For Finishing, Pure Transformer remained slightly above T-GAN after filtering, at 0.46 compared with 0.45. Before filtering, their scores were 0.47 and 0.45, respectively. The small difference provides little evidence of an additional graph-related benefit for this short, event-anchored phase. Using a 10-frame threshold preserved brief Finishing predictions, but filtering alone did not substantially improve their temporal agreement. Pure GNN reached 0.29 and the corrected rule-based model 0.36 after filtering.

Counter Attack showed the clearest T-GAN advantage over the neural baselines. T-GAN increased from 0.52 before filtering to 0.74 afterwards, compared with 0.44 to 0.60 for Pure Transformer. Pure GNN increased from 0.33 to 0.51, and the corrected rule-based model increased from 0.06 to 0.16. Random Forest increased from 0.23 before filtering to 0.39 afterwards. Counter Attacks often follow possession regain when defensive organisation is incomplete and temporary spatial imbalances emerge \citep{RN600,RN595}. Relational player representations may help distinguish these configurations from Progression, consistent with earlier work on player interactions in football \citep{RN602,RN473}.

Taken together, the comparisons support the value of temporal modelling for phase segmentation and an additional relational contribution to Counter Attack recognition. The benefit of graph information was smaller for broad intentions and Finishing. Filtering improved overall agreement, but its effects varied by class and did not replace the need for informative raw predictions.

\subsection{Typical Misalignments}

\begin{figure}[htbp]
    \centering
    \includegraphics[width=0.95\textwidth]{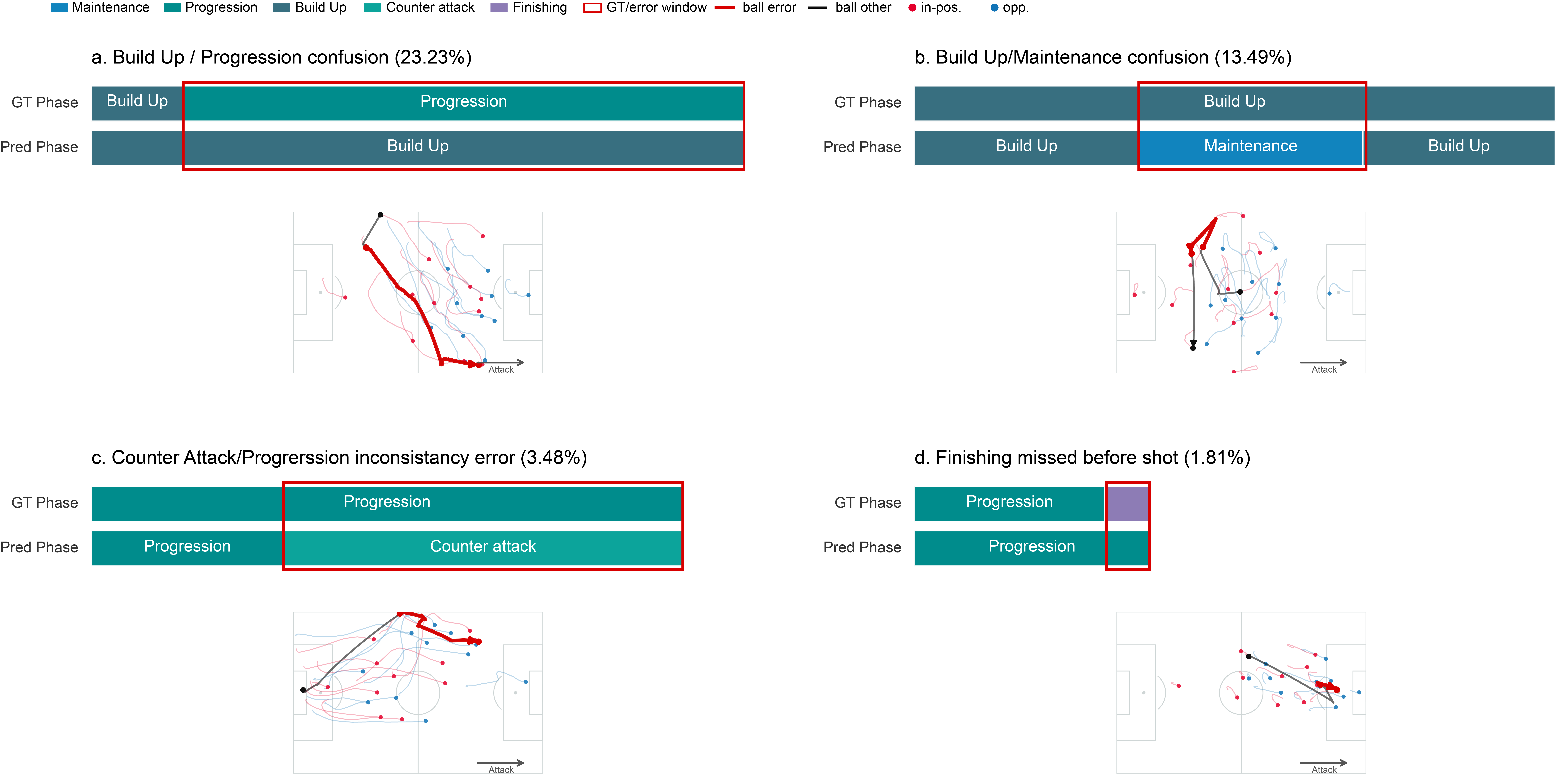}
    \caption{Representative sequence-level phase recognition misalignments. Misalignments were defined as ground-truth phase segments with tIoU < 0.5. The percentage in each title indicates the proportion of the corresponding type among all misalignments. The upper bars compare ground-truth and predicted phase sequences, and the red rectangles mark the mismatched windows. The pitch plots show the related ball trajectories, with red trajectories indicating the misalignment segments.}
    \label{figure6}
\end{figure}

To complement the updated aggregate comparison, representative misalignments from the original prediction set were retained for qualitative diagnosis. Figure~\ref{figure6} shows four examples, with misalignments defined as observed phase segments with tIoU below 0.5.

The most frequent misalignment was Build Up / Progression confusion, accounting for 23.23\% of all misalignments (Figure 6a). This misalignment occurs within the same high-level intention, Invade Opponent Space. In the example, the team progresses through a large diagonal switch and gradual forward development rather than a clearly separated vertical action. Such sequences may resemble Build Up because they involve circulation from deeper or wider areas, while their tactical effect is already progression \citep{RN625,RN624}. This suggests that the model sometimes struggles with diagonal-switch progression and gradual phase boundaries.

The second misalignment type was Build Up / Maintenance confusion, accounting for 13.49\% of all misalignments (Figure 6b). This reflects ambiguity between Invade Opponent Space and Keep Possession. Temporary backward or lateral passes under pressure may make an attacking construction sequence appear similar to ball retention, even when the broader intention remains to Invade Opponent Space.

The third misalignment type was Counter Attack / Progression inconsistency, accounting for 3.48\% of all misalignments (Figure 6c). Counter Attack and rapid Progression can share similar spatial and speed structures, such as fast forward movement and exploitation of open space. Although the model includes possession-regain information, longer temporal patterns may still be influenced by these similar movement structures.

The fourth misalignment type was missed Finishing before a shot, accounting for 1.81\% of all misalignments (Figure 6d). This mainly reflects the short and event-anchored nature of Finishing, where small boundary shifts can strongly reduce tIoU. This resembles a common issue in temporal action localisation, where precise start and end boundaries are often more difficult to recover than the presence of the action itself \citep{RN628}. Nevertheless, the action-biased Transformer still captured most shot-related cues: under a lenient criterion requiring only one correctly identified pre-shot Finishing frame, 79.80\% $\pm$ 12.49\% of such situations were recovered. Thus, the main issue is often the precise temporal extent rather than complete failure to detect Finishing.

Overall, these misalignments reflect ambiguity in intention, diagonal switching or gradual progression, similarity in transition-like movement, and short, event-anchored phases. They may partly relate to the limited sample size, as rare borderline situations are underrepresented in seven matches. After post-processing, however, these misalignments are unlikely to materially distort downstream analyses, which mainly rely on dominant tactical segments rather than exact frame-level boundaries. In applied use, deterministic post-processing can further improve the usability of the output by enforcing clear operational constraints, such as preventing implausible Counter Attack--Progression adjacencies and restoring short pre-shot Finishing segments. Such post-processing should be understood as a practical refinement step for tactical analysis, rather than a replacement for the data-driven model.

Future work could address these issues through larger annotated datasets, boundary-aware sequence models, stronger multi-scale context modelling, and more explicit representations of possession-regain context and opponent defensive organisation. For highly detailed case studies, uncertainty flags or manual review may still be useful for borderline sequences.

\subsection{Downstream Indicator Agreement}

Since the proposed framework is intended for practical performance analysis, we further examined whether automatically identified phases preserved the aggregate phase-specific information derived from manual annotations. Phase-resolved indicators were calculated from three label sources: Observed manual annotations, Identified labels after temporal filtering, and Fixed labels after deterministic correction. The indicators included phase distribution, mean phase duration, Player Speed, and Voronoi area at the team-average and Individual Ball Action (IBA) player levels. The fixed labels were obtained using two simple rules motivated by the typical misalignments: Progression and Counter Attack were not allowed to appear directly adjacent to each other, and if no Finishing label was detected within the ten frames before a shot inside the penalty area, these ten frames were reassigned to Finishing.

\begin{figure}[htbp]
    \centering
    \includegraphics[width=0.95\textwidth]{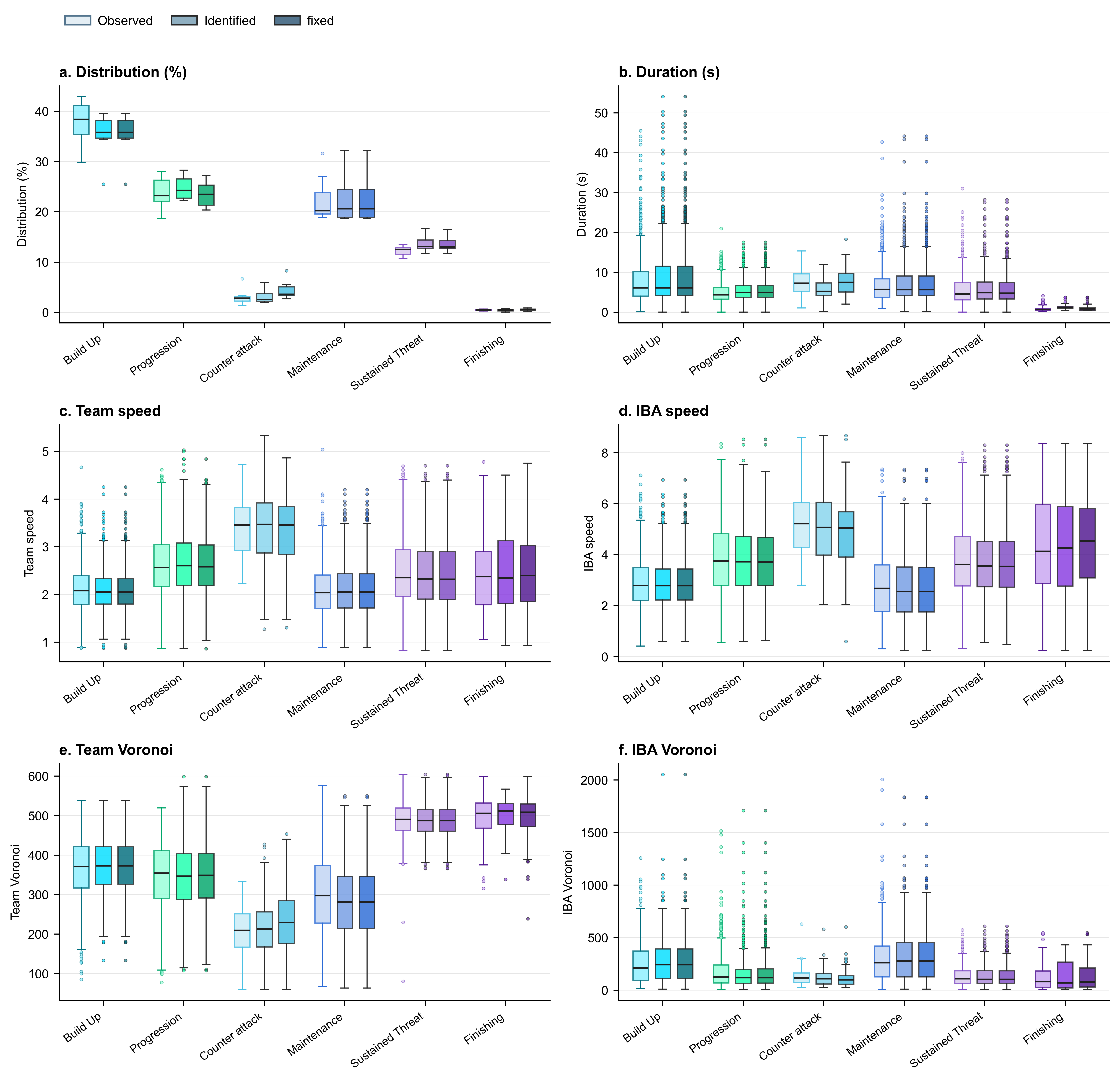}
    \caption{Agreement of phase-resolved indicators across Observed, Identified, and Fixed labels. Identified denotes the temporally filtered model output, and Fixed denotes the output after deterministic correction. Boxplots and points show distributions of phase structure, duration, movement, and spatial indicators across the six phases. The substantial overlap indicates that the automated labels preserved the aggregate profiles derived from manual annotations. IBA = Individual Ball Action.}
    \label{figure7}
\end{figure}

For Phases Distribution, paired Wilcoxon signed-rank tests and paired t-tests were used. For phase occurrence level indicators, where segments from different label sources could not be paired one-to-one, Mann--Whitney U tests and Welch t-tests were used. Holm correction was applied across the six phases for each indicator and comparison. No significant differences were found after Holm correction (See Figure~\ref{figure7}), indicating that the automated labels preserved the aggregate phase structure and phase-specific movement--spatial profiles of the manual annotations. Although Identified labels showed a visible decrease in Counter Attack duration, this distribution was restored after the fixing process. These results indicate that, for aggregate phase-based applications such as playing-style profiling, phase-specific movement analysis, and spatial occupation analysis, the automatically generated labels were practically comparable to manual annotations.

\section{Model Interpretation using Player Attention}

To improve the model's interpretability and examine whether its behaviour aligns with established football-specific patterns, we compute player attention weights that quantify each player group's (Forward, Midfielder, Winger, Defender, and Goalkeeper) relative contribution to the phase classification. Based on the GNN's learned representations, the model assigns higher attention to players whose behaviours are most informative for distinguishing tactical intentions and match phases. This mechanism allows us to identify (1). which positional groups the model relies on most strongly, and (2). whether these learned importance patterns are consistent with established football domain knowledge. By linking model decisions to specific player roles, player attention provides a transparent bridge between data-driven predictions and tactical interpretation.

\begin{figure}[htbp]
    \centering
    \includegraphics[width=0.95\textwidth]{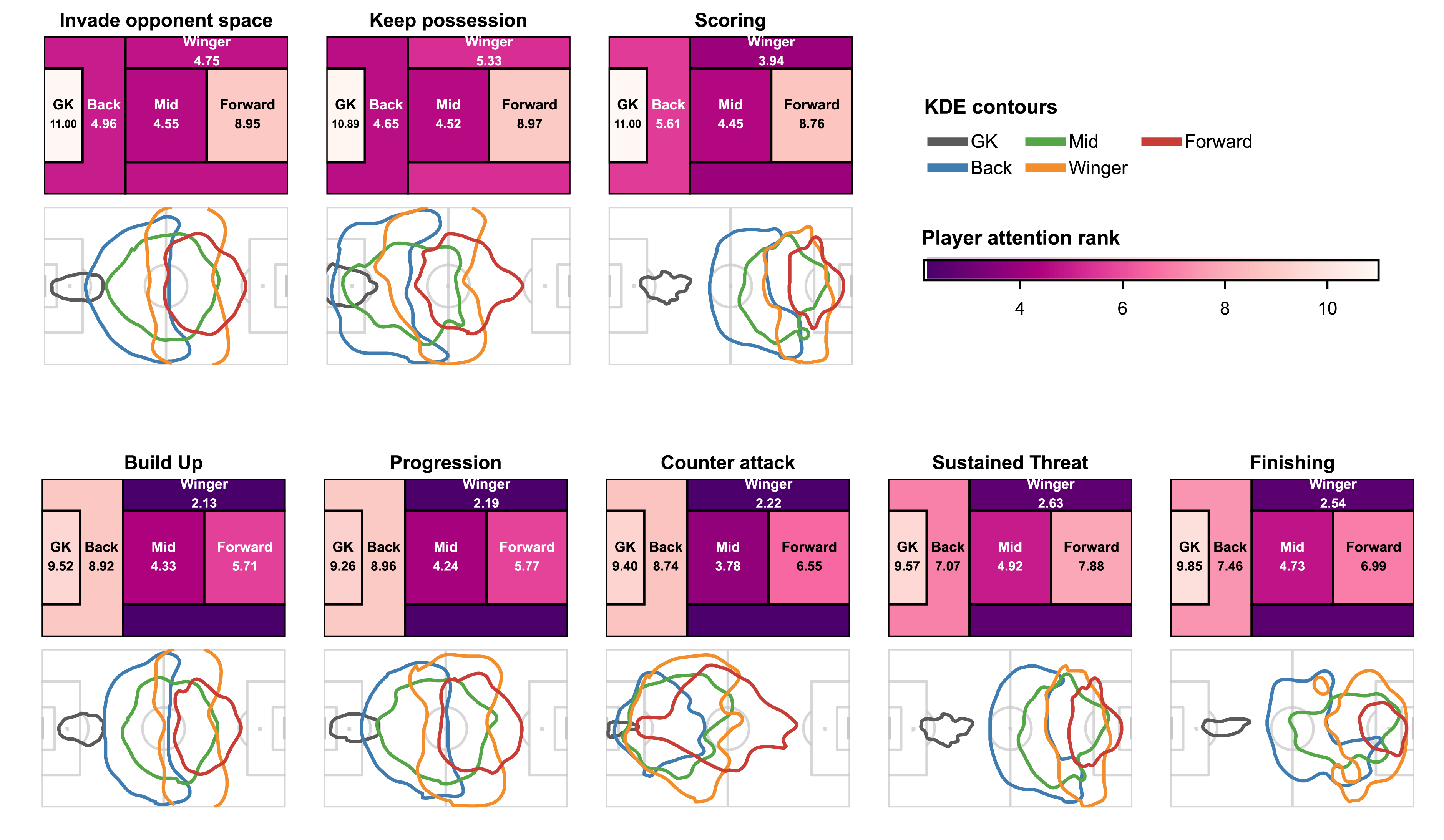}
    \caption{Group-level attention rankings and corresponding KDE-based spatial occupation maps for the three T-GAN models. Rows represent intention-level, invade-phase, and scoring-phase models, respectively. Numerical values denote attention rankings (lower = higher attention). KDE maps show the spatial distributions of positional groups for each class. (Mid=Midfielder, GK=Goalkeeper)}
    \label{figure8}
\end{figure}

Figure~\ref{figure8} summarises the learned group-level attention rankings for the three T-GAN models, together with the corresponding KDE-based spatial occupation maps. The results reveal distinct yet internally stable weighting strategies across the three models. Within each model, most positional groups maintain similar attention rankings across phases, with only minor variations.

At the intention level, the T-GAN consistently assigns the highest attention to Wingers, followed by Midfielders and Back players. Interestingly, under the Scoring intention, the ranking value of Wingers further decreases (corresponding to higher attention), representing the only noticeable deviation from strict stability within this model. This observation aligns with the well-established importance of pitch width in attacking organisation, where teams use wide positioning to stretch defensive structures and create exploitable spaces \citep{RN605,RN13}. By occupying wide areas of the pitch, wingers shape the horizontal structure of the attack and influence defensive compactness, a key element of offensive organisation in football. In this sense, the model appears to rely on the collective spatial configuration of the attacking unit rather than focusing solely on the most advanced central players.

In the invade opponent space model, Wingers also consistently receive the highest attention, followed by Midfielders and Forwards, while Back players and Goalkeepers rank substantially lower. The KDE maps reveal a clear spatial reconfiguration of Wingers across phases. During build-up, they mainly occupy a relatively narrow band in the front half of the pitch while maintaining lateral width. In Progression, their distribution extends more vertically, reflecting involvement across a broader attacking pathway. During Counter Attack, their activity shifts toward deeper starting positions and becomes more spatially dispersed. Midfielders also show a backward displacement in Counter Attack, likely reflecting transitional support and coverage, while Forwards display a more dispersed spatial pattern.

This suggests that when distinguishing invasion-intent phases, the model relies primarily on the behaviour of advanced, wide-attacking units to distinguish between Build Up, Progression, and Counter Attack phases. The KDE maps help illustrate the spatial mechanisms underlying these attention patterns. In particular, wide and advanced players exhibit clear longitudinal redistribution across phases, with counter-attacks showing deeper starting positions and broader spatial dispersion. Such changes again showed that the GNN part of the model captured the transitional nature of counterattacks, which typically occur when the defending team is temporarily disorganised and attacking players exploit open space before the defensive structure is restored \citep{RN600,RN595}.

In the scoring-phase model, Wingers and Midfielders remain the two most attended groups, with a smaller gap between them compared to the invade-phase model. Forwards receive only moderate attention, and their relative importance decreases compared with invasion-related phases. The KDE maps show subtle but consistent structural differences between Sustained Threat and Finishing. During Finishing, both Wingers and Midfielders exhibit a slight forward shift and spatial contraction, suggesting a more concentrated attacking configuration near the opponent's penalty area. Back players also show a mild narrowing, whereas Forward distributions remain relatively similar across the two phases.

This indicates that the model does not rely solely on the central striker's position when distinguishing between Sustained Threat and Finishing situations. Instead, it appears to capture coordinated structural adjustments across multiple attacking units. The slight forward shift and contraction of wide and midfield players during Finishing suggest that the model identifies this phase through collective attacking compression rather than through a single role-specific cue.

Overall, the player attention analysis suggests that T-GAN does not make phase predictions in an arbitrary way. Instead, its learned attention patterns are broadly consistent with football-specific tactical principles and provide an interpretable link between data-driven modelling and domain knowledge. This is particularly valuable in sports analytics, where the acceptance of machine learning models often depends not only on predictive performance but also on whether their outputs can be related to meaningful tactical knowledge. Comparable interpretability approaches, such as feature importance, have already been used in football analysis to connect model behaviour with domain-relevant variables \citep{RN168,RN50}. Although player attention should not be treated as a definitive tactical metric, it reflects the relative relevance of different players or positional groups to the model's phase discrimination process, improves the transparency of the proposed framework, and helps contextualise its predictions within interpretable spatial and positional structures.

\section{Practical Application}

To demonstrate the practical applications of the proposed framework, we analyse a Bundesliga match between FC Bayern Munich (FCB) and SC Paderborn (SCP) from the 2019--2020 season. Bayern Munich won the match 3--2 and largely controlled the game, holding approximately 70\% possession and producing 23 shots compared to Paderborn's 6, according to publicly available match statistics \citep{RN606}. Despite the clear territorial dominance, the scoreline remained relatively close, reflecting differences in attacking efficiency between the two teams. This type of possession-dominant scenario provides a useful context for illustrating how the proposed approach can be used in a specific game. In the following sections, we briefly demonstrate these applications from two perspectives:

phase-resolved sequence analysis of a tactically relevant match episode.

\begin{itemize}
\item Team-level playing style, reflected in phase distributions;
\item Phase driven sequence analysis of a tactically relevant match episode.
\end{itemize}

Beyond these examples, the framework could also potentially support several further practical applications. For instance, phase distributions may help examine match control and momentum by revealing how possession structures shift across different periods of a game. The framework could also be used to analyse tactical evolution within a match, such as structural changes following scoreline variations or substitutions. From a coaching perspective, the interpretable phase representation may support structured post-match feedback by providing a clearer description of how possession was organised, where attacking sequences broke down, or how turnovers developed into transition threats. In addition, the model may contribute to automated match annotation, assisting the large-scale labelling of possession sequences in tracking data analysis pipelines. However, such extended investigations are beyond the scope of the present study.

\subsection{Playing styles}

\begin{figure}[htbp]
    \centering
    \includegraphics[width=0.80\textwidth]{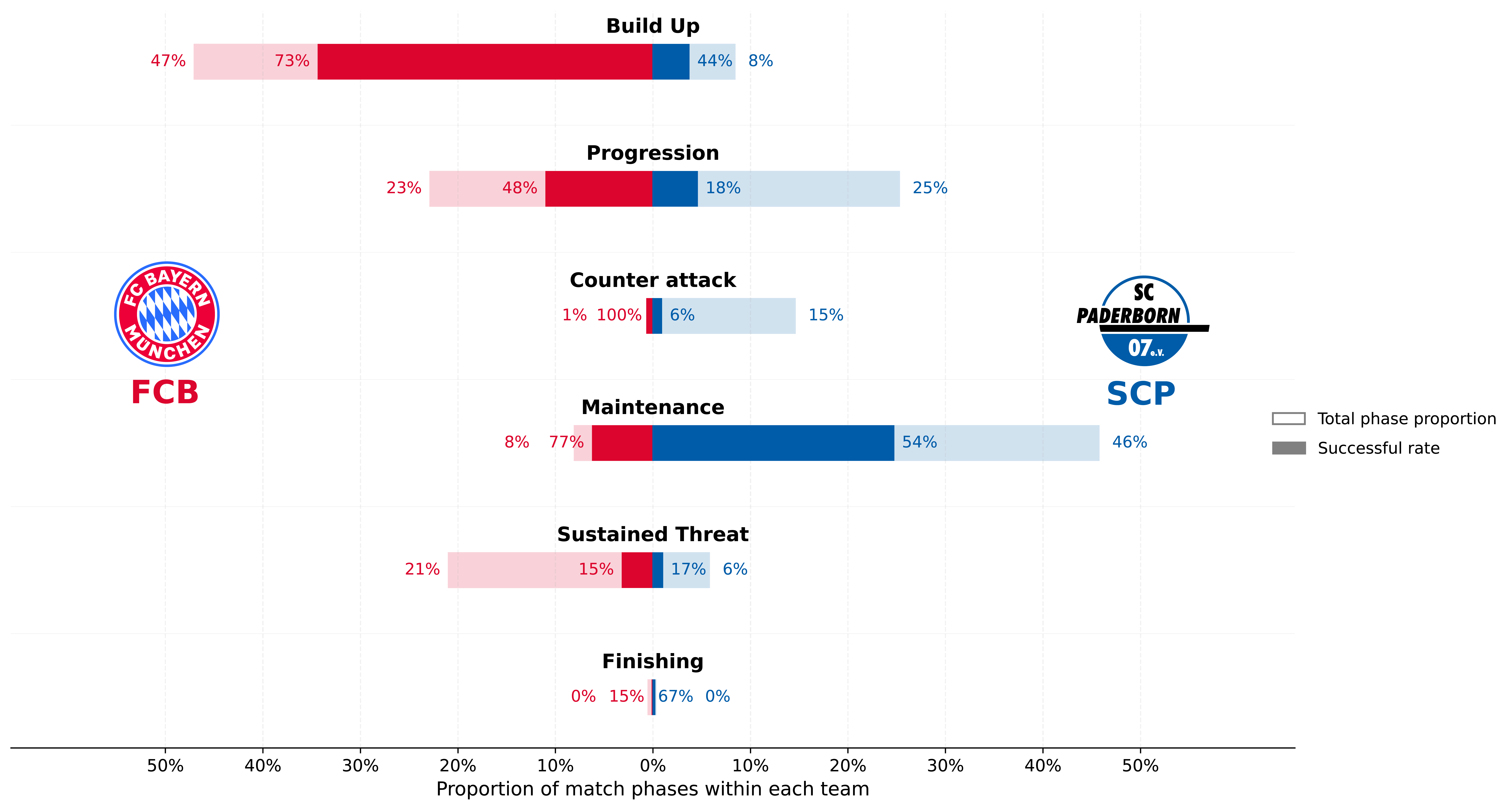}
    \caption{Distribution and success rates of match phases for FC Bayern Munich (FCB) and Paderborn (SCP) in the analysed Bundesliga match. Semi-transparent bars represent the proportion of each phase within the team's total possession time, while solid bars indicate the success rate of the corresponding phase transitions. For example, the success rate of the Build Up phase is the proportion of Build Up situations that progress to the Progression phase. Due to the very short duration of finishing situations, the phase proportion for Finishing is close to zero, and the success rate provides a more informative indicator of attacking efficiency.}
    \label{figure9}
\end{figure}

Figure~\ref{figure9} shows the phase distribution, which reveals a clear stylistic contrast between the two teams and reflects the match's overall dynamics.

FCB dominated possession and spent nearly half of their possession time in the Build Up phase (47\%), reflecting a structured midfield organisation and sustained control of the game. With a high Build Up success rate (73\%), FCB were able to progress the ball forward frequently, resulting in a considerable proportion of Progression phases (23\%). They also maintained prolonged pressure in the attacking third, with Sustained Threat accounting for 21\% of possession time. However, the efficiency of the final attacking stages was relatively low: only 15\% of sustained threats developed into finishing situations and only 15\% of those resulted in goals. This pattern is consistent with the match statistics, where FCB took 23 shots but scored only 3 goals.

In contrast, SCP's phase distribution reflects a team operating largely under pressure. Nearly half of their possession time occurred in the Maintenance phase (46\%), suggesting that much of their possession consisted of attempts to retain the ball under FCB's pressing rather than constructing organised attacks. Their Build Up proportion was very limited (8\%), indicating that FCB's high pressure frequently disrupted early-stage organisation. As a result, many of SCP's progression attempts occurred after temporary ball retention or through direct long passes, which may explain the relatively low progression success rate (18\%). Although Counter Attack phases accounted for 15\% of their possessions, the success rate was only 6\%, meaning that most transitions were interrupted before creating clear threats. Nevertheless, SCP were highly efficient when rare opportunities emerged, converting 67\% of finishing situations into goals (two goals from three shots on target).

Overall, the phase distribution captures the contrasting playing styles of the two teams: FCB exhibited a possession-dominant structure characterised by sustained Build Up and continuous territorial pressure, while SCP mainly relied on limited transition opportunities under heavy pressure. This case illustrates how the proposed match phase model can describe team playing styles in possession and partially reconstruct the flow of a match through interpretable phase distributions.

\subsection{Phase Driven Sequence Analysis}

\begin{figure}[htbp]
    \centering
    \includegraphics[width=0.95\textwidth]{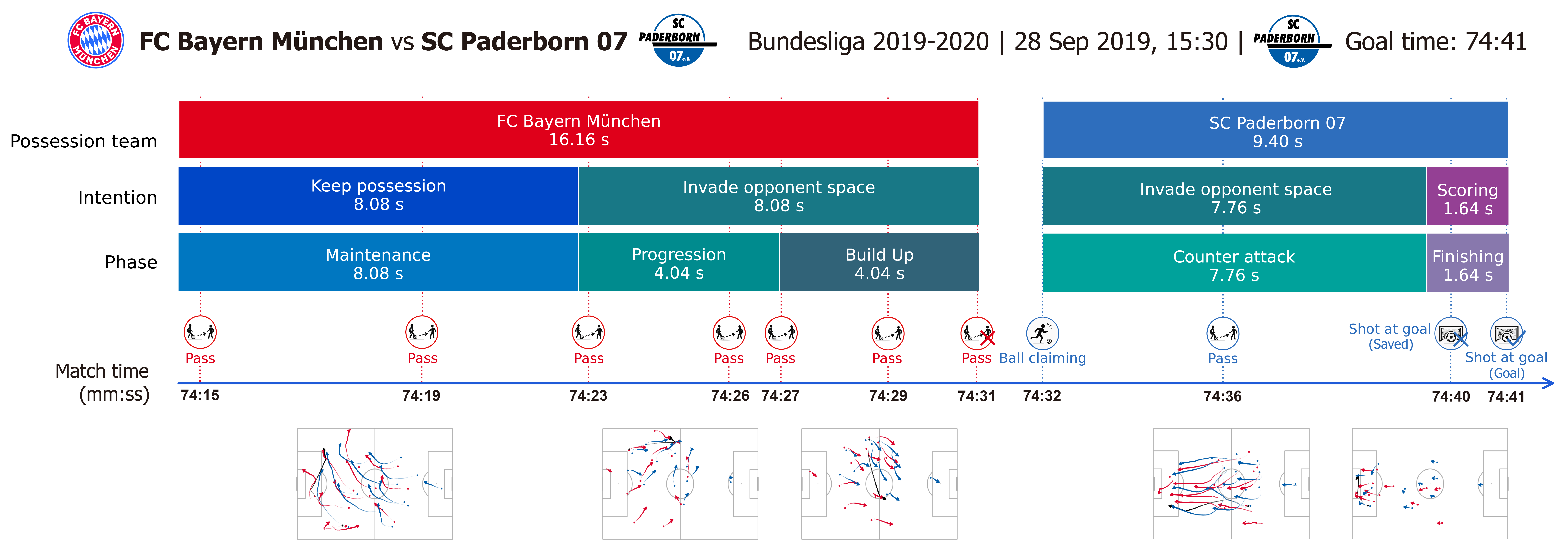}
    \caption{Phase-resolved timeline of the possession sequence preceding SC Paderborn's goal against FC Bayern Munich. The timeline combines possession episodes, tactical intentions, match phases, on-ball events, and selected tracking snapshots, illustrating how a longer FCB possession was followed by a rapid SCP Counter Attack and Finishing sequence.}
    \label{figure10}
\end{figure}

Figure~\ref{figure10} uses Paderborn's goal sequence as an example of phase-driven sequence analysis. Instead of describing the episode only as a chain of isolated events, such as passes, ball recovery, shots, and the goal itself, the timeline reconstructs how possession, tactical intention, match phase, and on-ball actions evolved together before the decisive moment.

The sequence first shows a 16.16s FCB possession. FCB initially remained in Keep Possession/Maintenance, indicating a period of controlled ball retention, before shifting into Invade Opponent Space through Progression and Build Up. However, this possession did not develop into a Scoring intention, suggesting that territorial control was not converted into a direct attacking threat. After the change of possession, SCP's attack developed in a much shorter but more decisive manner. Following ball claiming, SCP immediately entered the Invade Opponent Space intention and was classified as Counter Attack for 7.76 s, before moving into a short Finishing phase that resulted in the goal.

This example links the aggregate playing-style analysis with a concrete match episode. While Bayern showed a possession-dominant structure at the match level, this sequence highlights how such dominance can still be vulnerable when possession is lost after a single passing error, and the opponent can exploit the resulting temporary defensive imbalance. Conversely, SCP's short possession illustrates how a team with limited overall possession can create high-impact attacking moments through rapid transition. For analysts and coaches, such timelines provide a structured basis for reviewing phase transitions, defensive transition problems, failed progressions, and efficient attacking sequences. Although the example focuses on the possession sequence preceding SCP's goal, the same approach can be applied to other tactically relevant episodes, such as turnovers, failed Counter Attacks, long attacking possessions, sequences before shots, or periods of sustained pressure.

\section{Limitations \& Future Works}

Despite the promising results of the proposed framework, several limitations should be acknowledged and serve as directions for future research.

First, the match phase annotations were conducted by a single researcher rather than through multi-annotator consensus or expert-coaching validation. Although the labelling process followed clearly defined rules and operational definitions to ensure internal consistency, the absence of multiple annotators prevents the assessment of inter-rater reliability. Future research could address this issue by involving multiple analysts or professional coaches in the annotation process, thereby enabling evaluation of annotation agreement and further validating the proposed phase definitions based on intention.

Second, the dataset used in this study contains only seven matches. While the leave-one-match-out evaluation provides a match-level validation strategy, the limited dataset may still influence model comparisons, the generalisability of player-attention patterns, and some typical misalignment patterns. Rare or borderline situations, such as diagonal-switch progression and short pre-shot finishing phases, may be underrepresented in only seven matches.

Third, the player attention analysis remains preliminary. The attention weights indicate which positional groups the model relies on for phase recognition, but they should not be interpreted as direct measures of tactical influence or player performance. Future work should test the stability of these patterns across matches, teams, and tactical contexts, and validate them against external football knowledge or independent performance indicators.

Fourth, the practical application was demonstrated only through two illustrative examples from a single match. Although these examples suggest potential value for tactical interpretation, they do not yet provide a systematic evaluation of their applied use in performance analysis. Future work should extend the framework to larger-scale analyses of playing style, phase efficiency, phase stability, and phase transition patterns. For example, phase distributions could be compared across teams, opponents, home--away contexts, and match outcomes, while phase-based indicators could quantify how quickly teams progress between phases, how stable each phase is, or which phase-sequence archetypes occur most frequently. Such analyses would help translate automated phase classification into coach-relevant diagnostics and broader sports-science investigations of tactical behaviour.

Finally, the present study focuses exclusively on in-possession phases due to their relatively clear tactical structure and stronger intentional characteristics. Extending the framework to include out-of-possession phases represents an important direction for future research and would enable a more comprehensive representation of match dynamics.

\section{Conclusion}

This study proposed a data-driven framework for identifying in-possession match phases in football based on their underlying tactical intentions. By organising match phases within an intention-based conceptual structure and modelling possessions as temporally evolving player-interaction graphs, the proposed approach integrates relational and sequential information from tracking data to infer phase boundaries and tactical behaviours.

The model comparisons support the value of temporal modelling for phase recognition and an additional relational contribution to Counter Attack. Temporal Intersection over Union complements frame-level evaluation by measuring agreement between observed and predicted segments. The original representative examples illustrate intention ambiguity, gradual progression, counterattack-like movements, and short Finishing phases. Player attention analysis further provides an exploratory link between predictions and positional structures.

Taken together, the proposed framework provides a computational approach for translating continuous tracking data into tactically meaningful representations of possession dynamics. Such representations may support large-scale tactical analysis, playing style profiling, and automated match annotation in football analytics.

\section*{Funding Details}

Yuesen Li is funded by the China Scholarship Council (No. 202206520002) and the Munich Data Science Institute (MDSI) at Technical University of Munich (TUM) via the MDSI Kickstarter Seed Funds program (ASPAS). The publication of the study was supported by the Technical University of Munich (TUM, www.tum.de) in the framework of the Open Access Publishing Program. These institutions had no role in study design, data collection and analysis, decision to publish, or manuscript preparation.

\section*{Ethics Approval}

Since each player agreed to data recording during matches upon signing their player license, special ethics committee approval for this study was not required. Nevertheless, all procedures performed in the study were in strict accordance with the Declaration of Helsinki and the ethical standards of the local ethics committee.

\bibliographystyle{surnameinitialnat}
\bibliography{litrature}

\clearpage
\section*{Appendix}
\setcounter{table}{0}
\renewcommand{\thetable}{A\arabic{table}}

\subsection*{Full Feature Set}
\label{app:full-feature-set}

\begin{table}[!htbp]
\captionsetup{font=scriptsize,labelfont=bf}
\centering
\caption{Input features used in the T-GAN model. Features are grouped into node-level player features, edge-level interaction features, and frame-level global contextual features.}
\label{tab:appendix-feature-set}

\scriptsize
\setlength{\tabcolsep}{1.5pt}
\renewcommand{\arraystretch}{0.86}

\begin{tabularx}{\textwidth}{@{}p{0.9cm} p{1.7cm} p{2.3cm} X p{1.25cm}@{}}
\toprule
\textbf{Level} & \textbf{Group} & \textbf{Feature} & \textbf{Definition} & \textbf{Data Type} \\
\midrule

\multirow[t]{11}{*}{Node}
& \multirow[t]{11}{*}{--}
& Player x & X coordinate of the player on the pitch (m) & Continuous \\
& & Player y & Y coordinate of the player on the pitch (m) & Continuous \\
& & Player v & Player velocity (m/s) & Continuous \\
& & Player a & Player acceleration (m/s$^2$) & Continuous \\
& & Player angle & Angle between the player moving vector and the player-goal vector ($^\circ$) & Continuous \\
& & Player-Ball distance & Distance between player and ball (m) & Continuous \\
& & Player-Ball angle & Angle between the player moving vector and the ball moving vector ($^\circ$) & Continuous \\
& & Player-Ball relative speed & Relative speed between the player and the ball (m/s) & Continuous \\
& & Player Voronoi area & Player's spatial control area via Voronoi tessellation (m$^2$) & Continuous \\
& & Is team in possession & Whether the player's team is in possession & Bool \\
& & Is IBA player & Whether the player is involved in an Individual Ball Action & Bool \\

\midrule

\multirow[t]{3}{*}{Edge}
& \multirow[t]{3}{*}{--}
& If teammate & Whether the two connected players are teammates & Bool \\
& & Player distance & Distance between the two connected players (m) & Continuous \\
& & Player angle & Angle between the movement vectors of the two connected players ($^\circ$) & Continuous \\

\midrule

\multirow[t]{26}{*}{Global}
& \multirow[t]{14}{1.7cm}{Spatial Context}
& Ball x & X coordinate of the ball on the pitch (m) & Continuous \\
& & Ball y & Y coordinate of the ball on the pitch (m) & Continuous \\
& & Ball z & Z coordinate of the ball on the pitch (m) & Continuous \\
& & Ball v & Ball velocity (m/s) & Continuous \\
& & Ball a & Ball acceleration (m/s$^2$) & Continuous \\
& & Ball angle & Angle between the ball moving vector and the ball-goal vector ($^\circ$) & Continuous \\
& & Ball-goalline angle & Angle between the ball moving vector and the opponent's goal line ($^\circ$) & Continuous \\
& & X distance team convex center to goal & X distance from the team's convex hull center to the opponent's goal (m) & Continuous \\
& & X distance opponent convex center to own goal & X distance from the opponent's convex hull center to the team's own goal (m) & Continuous \\
& & Team average speed & Average speed of all outfield players from the team in possession (m/s) & Continuous \\
& & Opponents behind ball & Number of opponents behind the ball & Continuous \\
& & Last defender x position & X coordinate of the opponent's last defender (m) & Continuous \\
& & X distance ball to last defender & X distance from the ball to the last defender (m) & Continuous \\
& & Distance ball to closest opponent & Distance from the ball to the closest opponent (m) & Continuous \\

\cmidrule(lr){2-5}

& \multirow[t]{9}{1.7cm}{Action}
& Action type & Type of the current action & One-hot \\
& & Is Possession Team & Whether the action is made by the possession team & Bool \\
& & Is Successful & Whether the action is successful & Bool \\
& & Start x & Starting x coordinate of the action (m) & Continuous \\
& & Start y & Starting y coordinate of the action (m) & Continuous \\
& & End x & Ending x coordinate of the action (m) & Continuous \\
& & End y & Ending y coordinate of the action (m) & Continuous \\
& & Delta x & Difference between End x and Start x (m) & Continuous \\
& & Action-goal angle & Angle between the action moving vector and the action starting point-goal vector ($^\circ$) & Continuous \\

\cmidrule(lr){2-5}

& \multirow[t]{3}{1.7cm}{Possession Context}
& Last possession opponent & Whether the last possession belongs to the opponent of the current possession team & Bool \\
& & Last possession length & Length of the last possession, measured by number of frames & Continuous \\
& & Possession gap & Gap between the current possession and the last possession, measured by number of frames & Continuous \\

\bottomrule
\end{tabularx}

\vspace{0.5em}
\begin{minipage}{0.95\textwidth}
\tiny
\textit{Note.} IBA = Individual Ball Action. It begins when a player is able to perform an action with the ball and had no IBA prior to this, and ends when the player is unable to perform any further action with the ball.
\end{minipage}

\end{table}

\clearpage
\subsection*{Training Strategy}
\label{app:training-strategy}

\begin{table}[!htbp]
\centering
\captionsetup{font=scriptsize,labelfont=bf}
\caption{The setting-ups of the T-GAN and each baseline model.}
\label{tab:appendix-training-strategy}

\scriptsize
\setlength{\tabcolsep}{3pt}
\renewcommand{\arraystretch}{2.05}

\begin{tabularx}{\textwidth}{@{}p{3.0cm} p{1.3cm} p{3.0cm} X@{}}
\toprule
\textbf{Model} & \textbf{Level} & \textbf{Features} & \textbf{Parameters and Setting-ups} \\
\midrule

\textbf{Rule-based model}
& Frame
& --
& \textbf{Input:} every frame. Intentions are not considered.
When the ball is moving forward, \textbf{Counter attack} is identified if:
(1) the last possession sequence belongs to the opponent;
and (2) the gap between the last and current possession sequence is less than 25 frames.
\textbf{Progression} is identified when the requirements of Counter attack are not met.
When the ball is not moving forward, \textbf{Build Up} is identified when the ball is in the middle third;
\textbf{Maintenance} when the ball is in the back third;
\textbf{Sustained Threat} when the ball is in the final third;
and \textbf{Finishing} when the ball is in the opponent penalty box. \\

\midrule

\textbf{Random Forest}
& Frame
& Global features
& \textbf{Input:} every frame.
\textbf{Parameters:} learning rate = 0.01, max depth = 4, number of estimators = 200, number of leaves = 10. \\

\midrule

\textbf{Pure GNN}
& Frame
& Graph features; global features input directly to GNN
& \multirow[t]{3}{\linewidth}{
\textbf{Input:} every stable ball possession or intention sequence, shuffled.
Batch size = 16; number of epochs = 40.
\textbf{Focal loss weight:}
Intention classifier: Invade opponent space = 0.2, Keep possession = 0.8, Scoring = 1.2;
Invade opponent space classifier: Build Up = 0.1, Progression = 0.2, Counter attack = 0.4;
Scoring classifier: Sustained Threat = 0.1, Finishing = 0.8.
} \\

\textbf{Pure Transformer}
& Sequence
& Global features
& \\

\textbf{T-GAN}
& Sequence
& Graph features; global features
& \\

\bottomrule
\end{tabularx}

\end{table}

\newpage
\subsection*{T-GAN Model}
\label{app:t-gan-model}

\subsubsection*{Frame-level Player Graph Encoding}
\label{app:frame-level-player-graph-encoding}

For each frame $t$, the players on the pitch are represented as a graph $G_t$, where nodes correspond to players and edges encode player--player interactions. After graph construction, a graph neural network (GNN) is applied to learn player representations from the frame-level interaction structure. The resulting embedding of player $i$ at frame $t$ is denoted as $\mathbf{h}_{t,i}$. This process can be written as:
\begin{equation}
\mathbf{H}_t = \operatorname{GNN}(G_t).
\end{equation}

where
\begin{equation}
\mathbf{H}_t = \left\{\mathbf{h}_{t,1},\mathbf{h}_{t,2},\ldots,\mathbf{h}_{t,N_t}\right\},
\end{equation}
and $N_t$ is the number of valid player nodes in frame $t$. In this way, each player embedding captures not only the player's own state, but also relational information derived from surrounding teammates and opponents.

\subsubsection*{Role-conditioned Player Attention}
\label{app:role-conditioned-player-attention}

To aggregate multiple player embeddings into a single frame-level representation, we introduce a player attention pooling mechanism. Each player is assigned a positional role label (e.g., goalkeeper, defender, midfielder, or forward), represented by a discrete variable $r_i$. A learnable role embedding $\mathbf{e}_{r_i}$ is then associated with each player role. The fused player embedding and the corresponding role embedding are concatenated and passed to an attention scoring network:
\begin{equation}
s_{t,i} = \operatorname{MLP}_{\mathrm{att}}\left(\left[\mathbf{h}_{t,i} \parallel \mathbf{e}_{r_i}\right]\right).
\end{equation}
where $[\bullet \parallel \bullet]$ denotes vector concatenation and $s_{t,i}$ is the unnormalised importance score of player $i$ at frame $t$. This design allows the attention mechanism to take into account both the player's current contextualised state and their positional role when estimating importance.

\subsubsection*{Team-normalised Attention Weights}
\label{app:team-normalised-attention-weights}

Because offensive and defensive players may contribute differently to phase recognition, and because the number of valid players in each group may vary, attention scores are normalised separately within each team group. Let $c_i \in \{0,1\}$ denote the team indicator of player $i$, where $c_i=1$ represents the team in possession and $c_i=0$ represents the defending team. The normalised attention weight of player $i$ is then computed as:
\begin{equation}
\alpha_{t,i} = \frac{\exp(s_{t,i})}{\sum_{j \in \mathcal{G}(c_i)} \exp(s_{t,j})}.
\end{equation}
where $\mathcal{G}(c_i)$ denotes the set of valid players belonging to the same team group as player $i$. This team-wise normalisation ensures that player importance is evaluated relative to players on the same team, rather than across all players. As a result, the contribution of a key defender is not diluted simply because more attacking players are involved in a given frame.

If a player node has missing team or role information, it is excluded from the normalisation and assigned zero attention weight. This masking strategy prevents invalid nodes from influencing the pooled frame representation.

\subsubsection*{Attention-based Frame Representation}
\label{app:attention-based-frame-representation}

After obtaining the normalised player attention weights, the frame-level representation is computed as the weighted sum of the fused player embeddings:
\begin{equation}
\mathbf{f}_t = \sum_{i=1}^{N_t} \alpha_{t,i}\mathbf{h}_{t,i}.
\end{equation}
The vector $\mathbf{f}_t$ summarises the player interaction structure of frame $t$, while giving greater influence to players assigned higher importance weights. In contrast to simple mean pooling, this attention-based aggregation enables the model to focus on players most informative for phase recognition in the current frame.

From an interpretability perspective, the attention weights $\alpha_{t,i}$ can also be directly inspected to identify which players contributed most strongly to the frame representation. This provides a useful bridge between the learned model representation and tactical analysis.

\subsubsection*{Action-conditioned Temporal Encoding}
\label{app:action-conditioned-temporal-encoding}

After obtaining the frame-level representation $\mathbf{f}_t$ from player-attention pooling, we further incorporate event-level information through an action embedding. For each frame $t$, the discrete action type $a_t^{\mathrm{type}}$ is first mapped to a learnable embedding, then concatenated with the corresponding continuous action features $\mathbf{a}_t^{\mathrm{cont}}$. The resulting vector is projected by a multilayer perceptron:
\begin{equation}
\mathbf{e}_t = \operatorname{MLP}_{\mathrm{act}}\left(\left[\operatorname{Emb}\left(a_t^{\mathrm{type}}\right) \parallel \mathbf{a}_t^{\mathrm{cont}}\right]\right).
\end{equation}
The graph-based frame embedding and action embedding are then concatenated and projected into the Transformer input space:
\begin{equation}
\widetilde{\mathbf{f}}_t = \mathbf{W}_p\left[\mathbf{f}_t \parallel \mathbf{e}_t\right] + \mathbf{b}_p.
\end{equation}
where $\mathbf{W}_p$ and $\mathbf{b}_p$ are learnable projection parameters. A positional encoding $\mathbf{p}_t$ is then added to preserve the temporal order of frames:
\begin{equation}
\mathbf{z}_t^{(0)} = \widetilde{\mathbf{f}}_t + \mathbf{p}_t.
\end{equation}

\subsubsection*{Event Gating}
\label{app:event-gating}

To strengthen the representation of frames containing explicit events, we apply an event gate before the Transformer encoder. Let $m_t \in \{0,1\}$ indicate whether frame $t$ contains an event. A gating vector is computed from the action embedding:
\begin{equation}
\mathbf{g}_t = \operatorname{Sigmoid}\left(\operatorname{MLP}_{\mathrm{gate}}\left(\mathbf{e}_t\right)\right).
\end{equation}
The gated frame representation is then:
\begin{equation}
\widehat{\mathbf{z}}_t^{(0)} = \mathbf{z}_t^{(0)} + m_t\left(\mathbf{g}_t \odot \mathbf{z}_t^{(0)}\right).
\end{equation}
where $\odot$ denotes element-wise multiplication. Thus, only event-bearing frames are selectively amplified.

\subsubsection*{Event-biased Self-attention}
\label{app:event-biased-self-attention}

Temporal dependencies across frames are modelled by an event-biased self-attention mechanism. For each frame $t$, an event-dependent bias term is computed from its action embedding:
\begin{equation}
b_t = m_t \cdot \operatorname{MLP}_{\mathrm{bias}}\left(\mathbf{e}_t\right).
\end{equation}
This bias is added to the attention score of frame $t$, so that frames associated with salient actions become more likely to be attended to across the sequence. The unnormalised attention score can therefore be written as:
\begin{equation}
s_{t,r} = \operatorname{Attn}\left(\widehat{\mathbf{z}}_t^{(0)},\widehat{\mathbf{z}}_r^{(0)}\right) + b_r.
\end{equation}
where $s_{t,r}$ is the unnormalised attention score from the current frame $t$ to the attended frame $r$, and $b_r$ is the event-dependent bias for frame $r$.

Then the normalised attention weight is obtained by a softmax:
\begin{equation}
\alpha_{t,r} = \frac{\exp(s_{t,r})}{\sum_{r'=1}^{T}\exp(s_{t,r'})}.
\end{equation}
where $T$ is the length of the current possession sequence.

\subsubsection*{Frame-wise Prediction}
\label{app:frame-wise-prediction}

Each Transformer layer follows a residual structure with layer normalisation and a position-wise feed-forward network. After the final layer, the contextualised representation of each frame is retained:
\begin{equation}
\mathbf{z}_1^{(l)},\mathbf{z}_2^{(l)},\ldots,\mathbf{z}_T^{(l)}.
\end{equation}
Then, each frame representation $\mathbf{z}_t^{(l)}$ is independently passed to the prediction head:
\begin{equation}
y_t = \operatorname{MLP}_{\mathrm{cls}}\left(\mathbf{z}_t^{(l)}\right).
\end{equation}
For multi-class tasks, a softmax function is applied to obtain class probabilities for each frame, whereas for binary tasks a sigmoid function is used. In this way, the Transformer serves to contextualise every frame using information from the full possession sequence, while the final prediction remains frame-wise. Thus, each frame label is informed not only by its local spatial configuration, but also by the surrounding temporal and event context.

\end{document}